\documentclass[sigconf]{acmart}
\renewcommand\footnotetextcopyrightpermission[1]{} % removes footnote with conference information in first column
\usepackage[normalem]{ulem}
\usepackage{amsmath}
\usepackage{enumitem}
\usepackage[ruled,linesnumbered]{algorithm2e}

\newtheorem{Definition}{Definition}

\AtBeginDocument{%
  }

\setcopyright{acmcopyright}
\copyrightyear{2023}
\acmYear{2023}
\acmDOI{XXXXXXX.XXXXXXX}

\acmConference[Conference acronym 'XX]{Make sure to enter the correct
  conference title from your rights confirmation emai}{June 03--05,
  2023}{Woodstock, NY}
\acmPrice{15.00}
\acmISBN{978-1-4503-XXXX-X/18/06}

\begin{document}

\title{Class-aware and Augmentation-free Contrastive Learning from Label Proportion}

\author{Jialiang Wang$^1$, Ning Zhang$^3$, Shimin Di$^1$, Ruidong Wang$^4$, Lei Chen$^{2,1}$}
\affiliation{
  \institution{$^1$The Hong Kong University of Science and Technology, Hong Kong SAR, China}
  \city{$^2$The Hong Kong University of Science and Technology (Guangzhou), Guangzhou, China \\}
  \state{$^3$Tencent America, Palo Alto, CA, USA \\}
  \country{$^4$Tencent Europe B.V., Amsterdam, North Holland, Netherlands}}
\email{{jwangic, sdiaa}@connect.ust.hk, {drningzhang, ruidwang}@global.tencent.com, leichen@cse.ust.hk}
\orcid{0009-0005-0850-0389}

% \author{Jialiang Wang}
% \affiliation{%
%   \institution{Hong Kong University of Science and Technology}
%   \city{Hong Kong SAR}
% \email{jwangic@connect.ust.hk}
% \orcid{0009-0005-0850-0389}

% \author{Ning Zhang}
% \affiliation{%
% 	\institution{Tencent America}
% 	\city{Palo Alto}
% 	\state{CA}
% 	\country{USA}}
% \email{drningzhang@global.tencent.com}

% \author{Shimin Di}
% \affiliation{%
%   \institution{Hong Kong University of Science and Technology}
%   \city{Hong Kong SAR}
% \email{sdiaa@connect.ust.hk}
% \orcid{0000-0002-7394-0082}

% \author{Ruidong Wang}
% \affiliation{%
% 	\institution{Tencent Europe B.V.}
% 	\city{Amsterdam}
% 	\state{North Holland}
% 	\country{Netherlands}}
% \email{ruidwang@global.tencent.com}

% \author{Lei Chen}
% \affiliation{%
%   \institution{Hong Kong University of Science and Technology (Guangzhou), Hong Kong University of Science and Technology}
%   \city{Guangzhou}
% \email{leichen@cse.ust.hk}
% \orcid{0000-0002-8257-5806}

\renewcommand{\shortauthors}{Wang et al.}

\begin{abstract}
	Learning from Label Proportion (LLP) is a weakly supervised learning scenario in which training data is organized into predefined bags of instances, disclosing only the class label proportions per bag. This paradigm is essential for user modeling and personalization, where user privacy is paramount, offering insights into user preferences without revealing individual data. LLP faces a unique difficulty: the misalignment between bag-level supervision and the objective of instance-level prediction, primarily due to the inherent ambiguity in label proportion matching. Previous studies have demonstrated deep representation learning can generate auxiliary signals to promote the supervision level in the image domain. However, applying these techniques to tabular data presents significant challenges: 1) they rely heavily on label-invariant augmentation to establish multi-view, which is not feasible with the heterogeneous nature of tabular datasets, and 2) tabular datasets often lack sufficient semantics for perfect class distinction, making them prone to suboptimality caused by the inherent ambiguity of label proportion matching. To address these challenges, we propose an augmentation-free contrastive framework TabLLP-BDC that introduces class-aware supervision (explicitly aware of class differences) at the instance level. Our solution features a two-stage Bag Difference Contrastive (BDC) learning mechanism that establishes robust class-aware instance-level supervision by disassembling the nuance between bag label proportions, without relying on augmentations. Concurrently, our model presents a pioneering multi-task pretraining pipeline tailored for tabular-based LLP, capturing intrinsic tabular feature correlations in alignment with label proportion distribution. Extensive experiments demonstrate that TabLLP-BDC achieves state-of-the-art performance for LLP in the tabular domain.
\end{abstract}

%\begin{CCSXML}
%<ccs2012>
%  <concept_id>00000000.0000000.0000000</concept_id>
%  <concept_desc>Do Not Use This Code, Generate the Correct Terms for Your Paper</concept_desc>
%  <concept_significance>500</concept_significance>
%  <concept_id>00000000.00000000.00000000</concept_id>
%  <concept_desc>Do Not Use This Code, Generate the Correct Terms for Your Paper</concept_desc>
%  <concept_significance>300</concept_significance>
%  <concept_id>00000000.00000000.00000000</concept_id>
%  <concept_desc>Do Not Use This Code, Generate the Correct Terms for Your Paper</concept_desc>
%  <concept_significance>100</concept_significance>
%  <concept_id>00000000.00000000.00000000</concept_id>
%  <concept_desc>Do Not Use This Code, Generate the Correct Terms for Your Paper</concept_desc>
%  <concept_significance>100</concept_significance>
%</ccs2012>
%\end{CCSXML}

%\ccsdesc[500]{Do Not Use This Code~Generate the Correct Terms for Your Paper}
%\ccsdesc[300]{Do Not Use This Code~Generate the Correct Terms for Your Paper}
%\ccsdesc{Do Not Use This Code~Generate the Correct Terms for Your Paper}
%\ccsdesc[100]{Do Not Use This Code~Generate the Correct Terms for Your Paper}


\received{20 February 2023}
\received[revised]{12 March 2023}
\received[accepted]{5 June 2023}

\maketitle

\section{Introduction}
\label{sec:introduction}
Learning from Label Proportions (LLP) is a weakly supervised learning paradigm organizing training data into predefined bags of instances, where only the proportion of each class within these bags is known. Its primary goal is to develop an instance-level classifier capable of predicting labels for individual instances, overcoming the constraints of aggregated labels \cite{musicant2007supervised, quadrianto2008estimating}. This capability is crucial for tailoring personalized web experiences, enabling the derivation of meaningful insights for privacy-preserving content delivery and advertising strategies \cite{yuan2013real, bayer2020impact} from bag-level proportional labels shared in business collaborations \cite{o2022challenges}. Extensively applied in online advertising, examples include platforms like Google Ads \cite{GoogleAdsDataHub2023} and Facebook Ad Platform \cite{FacebookMetaBusinessHelpCenter2023}, alongside privacy-preserving strategies such as aggregated reporting and K-anonymity principles \cite{sweeney2002k}. LLP strikes a balance between enhancing user experience and maintaining privacy by leveraging aggregated data to reduce risks associated with detailed user profiles. Additionally, the cost-efficiency of storing aggregated data is particularly beneficial for large databases \cite{baeza2015predicting}. LLP offers a practical solution in scenarios where storing detailed user data is unnecessary, such as electoral campaigns \cite{qi2016learning, sun2017probabilistic} or demographic categorization \cite{ardehaly2017co}, allowing businesses to economically derive instance-level insights from aggregated data storage.

Advancing LLP methodologies necessitates addressing the inherent difficulty of aligning bag-level training objectives with instance-level predictions. Research in LLP has predominantly focused on the label proportion matching principles to train classifiers that accurately reflect observed label proportions \cite{zhang2022learning}. Traditional LLP models, such as SVMs \cite{felix2013psvm, qi2016learning}, Bayesian models \cite{hernandez2013learning}, and clustering techniques \cite{stolpe2011learning}, were mainly designed for binary classification and faced scalability issues. The advent of deep representation learning introduced a significant shift, leveraging deep neural models for extensive, multi-dimensional LLP predictions in the image domain \cite{ardehaly2017co, tsai2020learning, liu2021two, liu2022llp}. These models often utilize bag-level Cross-entropy \cite{dulac2019deep} loss to facilitate end-to-end learning in multiclass settings. However, a closer examination reveals a critical, yet often overlooked, difficulty in LLP: the misalignment between bag-level training objectives and instance-level predictions. This difficulty stems from the inherent ambiguity in label proportion matching, where multiple instance-level label distributions could feasibly correspond to identical bag-level label proportions \cite{la2022learning}. Although barely formally defined or theoretically analyzed, successful image-based LLP methods have employed instance-level supervision to address this challenge, as summarized in Tab.~\ref{tab:related_work_table}. These deep LLP models, tailored for image data, utilize methods such as adversarial autoencoders \cite{wang2023llp}, Self-supervised learning \cite{liu2022self}, and prototypical contrastive clustering \cite{la2022learning}, capitalizing on label-invariant augmentations like image rotation \cite{liu2022self} to achieve effective instance-level supervision without individual labels. Notably, contrastive-based models \cite{la2022learning} have demonstrated particular promise by generating instance-level contrastive signals between multiple views from single instances.

\begin{table*}[!t]
	\centering
	\caption{Comparison of existing works in Deep LLP.}
	\label{tab:related_work_table}
	\vspace{-10px}
	\begin{tabular}{|c|c|c|c|c|c|c|}
		\hline
		\multicolumn{1}{|c|}{\textbf{Deep LLP}} & \textbf{Scenario} & \textbf{Pretraining} & \textbf{Augmentation} & \multicolumn{3}{c|}{\textbf{Fine-tuning}} \\ \cline{2-7} 
		\multicolumn{1}{|c|}{\textbf{Strategy}} & Data Type & Task & \textbf{Technique} & Principle & \textbf{Ins. Sup.} & \textbf{\textit{Class-aware}} \\ 
		\hline
		DLLP \cite{ardehaly2017co} & Image/Text & pre-trained
		weights & - & Batch Averager & $\times$ & - \\ 
		\hline
		ROT \cite{dulac2019deep} & Image & pre-trained
		weights & label-invariant & Relax-OT Loss & \checkmark & \checkmark \\ 
		\hline
		LLP-VAT \cite{tsai2020learning} & Image & pre-trained
		weights & VAT & Consistency Regularization & \checkmark & $\times$ \\ 
		\hline
		LLP-PLOT \cite{liu2021two} & Image & pre-trained
		weights & mixup & OT-based Pseudo-labeling & \checkmark & \checkmark \\ 
		\hline
		SELF-LLP \cite{liu2022self} & Image & Self-supervised & label-invariant & Self-supervised \& -ensemble & \checkmark & \checkmark \\ 
		\hline
		LLP-GAN \cite{liu2022llp} & Image/Text & - & label-invariant & GAN & \checkmark & $\times$ \\ 
		\hline
		LLP-AAE \cite{wang2023llp} & Image & - & label-invariant & Adversarial Autoencoder & \checkmark & $\times$ \\ 
		\hline
		LLP-Co \cite{la2022learning} & Image & pre-trained
		weights & label-invariant & Prototypical Contrastive & \checkmark & \checkmark \\ 
		\hline
		SelfCLR-LLP \cite{nandy2022domain} & Tabular & - & free & Self-contrastive & \checkmark & $\times$ \\ 
		\hline
		TabLLP-BDC (Ours) & Tabular & Bag Contrastive & free & Bag Difference Contrastive & \checkmark & \checkmark \\ 
		\hline
	\end{tabular}
	\caption*{
		\small
		\textnormal{\textbf{Notations:} \textbf{pre-trained weights} indicate either the model explicitly uses a pretrained model on ImageNet \cite{deng2009imagenet} or adopts a backbone model for which a publicly available pretrained version exists. \textbf{label-invariant} means adopting label-invariant augmentations, such as image rotation, while \textbf{free} refers to an augmentation-free method. \textbf{Ins. Sup.} indicates the proposed method introduces instance-level supervision, and \textbf{\textit{Class-aware}} means the instance-level supervision is explicitly aware of class differences.}}
	\vspace{-20px}
\end{table*}

The urgency for applying LLP to tabular data has intensified, especially in sectors like e-commerce and online advertising, where balancing privacy with personalized analysis became crucial \cite{o2022challenges}. While native contrastive learning \cite{la2022learning} has proven effective in image-based LLP, it encounters significant challenges with tabular data. First, the heterogeneous and sensitive nature of tabular datasets compromises the feasibility of label-invariant augmentation techniques \cite{grinsztajn2022tree}, which are essential for generating effective positive and negative pairs or multi-view in native contrastive learning. This is because tabular data can be extremely fragile to tiny perturbations \cite{borisov2022deep}, especially when manipulating categorical features, thus highlighting the necessity for an augmentation-free contrastive approach. Moreover, the inherent inconsistency and absence of spatial interdependencies and semantics within tabular data diminish the robustness essential for perfect class distinction \cite{grinsztajn2022tree}. Such limitations exacerbate the suboptimality of relying exclusively on matching label proportions and auxiliary losses that lack explicit class distinction, or \textit{class-awareness}, at the instance level, intensifying the inherent difficulty of LLP. Consequently, there is a pronounced need for an augmentation-free and \textit{class-aware} contrastive-based LLP approach, which should be meticulously designed to accommodate the specific characteristics of tabular data. 

Recently, SelfCLR-LLP \cite{nandy2022domain} has pioneering work in tabular-based deep LLP, utilizing Self-contrastive auxiliary loss as an instance-level signal to foster diversity in representation without augmentation. However, as summarized in Tab.~\ref{tab:related_work_table}, this approach falls short of relying predominantly on a classical Self-supervised formulation without any incorporation of label proportions, thus failing to offer the crucial \textit{class-aware} guidance necessary for precise class distinction at the instance level. This limitation underscores a significant gap in achieving a tailored and effective contrastive solution for tabular-based LLP that can effectively utilize ambiguous label proportions for \textit{class-aware} instance-level supervision without the need for augmentation. Beyond this, no existing work proposes a pretraining pipeline tailored for tabular-based LLP, despite the promising prospects of LLP as a weakly supervised learning framework for pretraining endeavors.

To address these challenges, we present TabLLP-BDC, a two-stage Bag Difference Contrastive (BDC) framework crafted to effectively navigate the intricacies of tabular-based LLP through \textit{class-aware} and augmentation-free contrastive learning. Our Difference Contrastive fine-tuning technique harnesses label proportion differences between bags to establish augmentation-free instance-level supervision that is explicitly aware of class distinctions—facilitating similar representations for instances within the same class and divergent representations across classes, a notable challenge in the absence of explicit instance-level labels. Additionally, TabLLP-BDC introduces a pioneering augmentation-free and multi-task Bag Contrastive pretraining pipeline tailored for tabular-based LLP, capturing intrinsic tabular feature correlations in alignment with label proportion distribution. Our main contributions include:
\begin{itemize}[leftmargin=*,nosep]
	\item We delve into the realm of LLP within tabular data domains, identifying and addressing unique challenges that hinder the development of current deep LLP methodologies due to the inherent properties of tabular datasets.
	\item We introduce a novel bag contrastive learning framework tailored for tabular-based LLP that circumvents the need for label-invariant augmentations. By leveraging a Linear Sum Assignment Problem \cite{crouse2016implementing} to generate pseudo-positive pairs between bags with proportional labels, our approach facilitates \textit{class-aware} instance-level supervision without relying on explicit instance-level labels.
	\item We pioneer a multi-task pretraining phase specially designed for the tabular-based LLP to capture intrinsic tabular feature correlations that resonate with label proportion distribution. Incorporating a novel Bag Contrastive task in the metric learning manner at the bag level, our approach further refines the efficacy of native Self-contrastive pretraining pipeline \cite{somepalli2021saint} and ensures a seamless transition to downstream LLP training.
	\item We conducted thorough experiments on various public and real-world tabular datasets using different bagging strategies and bag sizes. Our analysis included both traditional LLP with instance-level validation and practical LLP with bag-level validation, introducing a novel evaluation metric, mPIoU, for the latter. Our TabLLP-BDC consistently delivers SOTA instance-level performance. Additionally, we conducted extensive ablation studies to assess each model component's impact and effectiveness.
\end{itemize}

\section{Related Work}
\subsection{Learning from Label Proportion}
Weakly Supervised learning (WSL) \cite{zhou2018brief, jiang2023weakly} emerges as a pragmatic alternative to fully supervised learning, especially in domains where obtaining dense and accurate labels is expensive or infeasible \cite{qi2016learning}. In its essence, WSL utilizes weak labels, which can be incomplete \cite{ruff2019deep, pang2018learning, tian2022anomaly}, inaccurate \cite{zhao2023admoe, dong2021isp, zhong2019graph}, or inexact \cite{sultani2018real, wan2020weakly, tian2021weakly}, to train machine learning models effectively \cite{zhou2018brief}. Among them, tasks using inexact labels without precise instance-level details have gained attention recently. A prominent example of such a task is Multiple Instance Learning (MIL) \cite{ilse2018attention, sultani2018real, wan2020weakly, tian2021weakly}, where models are trained on group-level labels and also make predictions at the group level.

However, there exists a particularly intriguing and less explored paradigm called Learning from Label Proportion (LLP). In the LLP framework, data is provided as bags of instances. A bag denoted as $B_k$, comprising $m$ instances. Each bag is associated with a label proportion vector $\mathbf{\bar{p}_k}$, where each entry $\bar{\mathbf{p}}_{k}^c = \frac{1}{m} \sum_{i=1}^{m} y_{ki}^c$ specifies the proportion of instances from a particular class $c$, assuming $\mathbf{y_{ki}}$ to be the potential one-hot label vector for instance $x_{ki}$ within the bag $k$. The LLP dataset can then be defined as $D_{LLP} = \left\{ (B_k, \mathbf{\bar{p}_k}) \right\}_{k=1}^{K}$ with $K$ being the total number of bags.

The uniqueness of LLP lies in its objective: while it is trained using group-level label proportions similar to MIL, it aims to make predictions at the instance level. Formally, the LLP problem is to learn an instance-level classifier $f_\theta(\cdot)$ parameterized by $\theta$ by minimizing an LLP loss function $\mathcal{L}_{LLP}(\cdot, \cdot)$, which aims to minimize the discrepancy between the predicted and actual label proportions $\mathbf{\bar{p}_k}$:
\begin{equation}
	\theta^* = \text{arg min}_\theta \sum_{k=1}^{K} \mathcal{L}_{LLP} \left(\frac{1}{m} \sum_{i=1}^{m} f_\theta(x_{ki}), \mathbf{\bar{p}_k} \right)
\end{equation}
where $\mathcal{L}_{LLP}(\cdot, \cdot)$ generally denotes a suitable discrepancy measurement. This formulation introduces the preliminary approach to LLP: ensuring that the aggregated instance-level predictions within a bag closely match the given bag label proportion. More advanced approaches necessitate the model to discern intricate patterns both within and across bags to infer likely labels for individual instances. 

\subsubsection{Traditional LLP}
The study of LLP in traditional machine learning began by reconfiguring supervised algorithms such as SVM, kNN, and Neural Network for aggregated outputs \cite{musicant2007supervised}, with typical adaptations including reassigning instance labels with bag label proportions. In the probabilistic realm, hierarchical Bayesian models trained with MCMC samplers \cite{kuck2012learning} or Structural EM \cite{hernandez2013learning} were introduced to address uncertainty. Other strategies included Mean Map models \cite{quadrianto2008estimating} that focus on mean operators for LLP optimization and clustering methods \cite{chen2009kernel, stolpe2011learning} that emphasize the intricate relationship between data structures and label proportions. The SVM landscape was later expanded with pseudo-labels \cite{felix2013psvm} and inverted conditional class probabilities estimation \cite{rueping2010svm}. This era concluded with the EPRM framework \cite{yu2014learning} that aimed to minimize empirical bag proportion loss. Despite these advancements, traditional LLP methods faced issues with scalability, rigid assumptions, and limited efficacy in multi-class contexts.
% The study of LLP in traditional machine learning began by reconfiguring supervised algorithms such as SVM, kNN, and Neural Network for aggregated outputs \cite{musicant2007supervised}, with typical adaptations including reassigning instance labels with bag label proportions. In the probabilistic realm, hierarchical Bayesian models trained with MCMC samplers \cite{kuck2012learning} were introduced to address uncertainty. Another Bayesian approach \cite{hernandez2013learning} used Structural EM for joint label assignments, moving beyond simple class label estimation. Additional strategies included Mean Map models \cite{quadrianto2008estimating} focusing on mean operators for LLP optimization, and clustering methods, such as the kernel K-means optimization \cite{chen2009kernel} tailored for aggregated outputs and cluster model optimization \cite{stolpe2011learning} that emphasizes the intricate relationship between data structures and label proportions. The SVM landscape was expanded with pseudo-labels \cite{felix2013psvm} and inverted conditional class probabilities estimation \cite{rueping2010svm}. This era concluded with the EPRM framework \cite{yu2014learning} that aimed to minimize empirical bag proportion loss. Despite these advancements, traditional LLP methods faced issues with scalability, rigid assumptions, and limited efficacy in multi-class contexts.

\subsubsection{Deep LLP}
With the rise of deep learning, LLP witnessed transformative approaches that redefined its landscape. Pioneering this shift, DLLP \cite{ardehaly2017co} employs deep networks to minimize the KL divergence between actual and predicted label proportions:
\begin{equation}
	\mathcal{L}_{\text{LLP}}(\mathbf{\hat{p_k}}, \mathbf{\bar{p_k}}) = \sum_{c=1}^{C} \mathbf{\bar{p_k}^c} \log  \frac{\mathbf{\bar{p_k}^c}}{\mathbf{\hat{p_k}^c}}
\end{equation}
where $\mathbf{\hat{p_k}=\frac{1}{m} \sum_{i=1}^{m} f_\theta(x_{ki})}$ is the batch-averaged classification probability, $\mathbf{\bar{p_k}}^c$ is the true bag label proportion of class $c$, $C$ is the number of class, $m$ is the bag size, $x_{ki}$ is the $i$th instance in the bag $k$, and $f_\theta(\cdot)$ is the instance-level classifier. This general LLP loss was adapted into a Cross-entropy form with an additional balanced Optimal Transport (OT) loss with entropic regularization by another study \cite{dulac2019deep}, signifying a move towards auxiliary loss functions and instance-level optimization for capturing label distributions \cite{liu2021two}. However, as these LLP models grappled with scaling to larger bag sizes, LLP-VAT \cite{tsai2020learning} attempted to mitigate this through consistency regularization and Virtual Adversarial Training \cite{miyato2018virtual}, but still encountered the suboptimal issue, underscoring the ongoing struggle for \textit{non-class-aware} supervision at the instance level. Diverging from these paths, LLP-GAN \cite{liu2022llp} championed the innovative GAN approach, melding generative adversarial techniques with LLP, a stride further propelled by LLP-AAE \cite{wang2023llp} through adversarial autoencoders. Meanwhile, models like LLP-Co \cite{la2022learning} combined Contrastive learning with OT to ensure label proportion-guided clustering consistency across multi-views, and SELF-LLP \cite{liu2022self} adopted Self-supervised techniques \cite{zhai2019s4l, gidaris2018unsupervised} alongside the Self-ensemble strategy \cite{laine2016temporal} to enhance representation robustness. Despite these methods have made significant advancements in the image, it reveals the gap in adapting these deep representation learning solutions to tabular data due to their reliance on image-specific techniques like label-invariant augmentation \cite{la2022learning, wang2023llp} or universal pre-training paradigm \cite{deng2009imagenet}. This underscores the need for domain-agnostic solutions in LLP, especially for larger bag sizes and tabular data.

Recently, SelfCLR-LLP \cite{nandy2022domain} introduced the first domain-agnostic contrastive learning framework aimed at the LLP challenge in tabular data. Leveraging an auxiliary Self-contrastive loss inspired by the NT-Xent loss \cite{chen2020simple}, this method aims to enhance diversity without augmentation. However, a critical limitation is that the auxiliary loss overlooks label proportions, rendering the approach devoid of \textit{class-awareness} at the instance level. Thus, it primarily enhances general-purpose representation diversity rather than providing a bespoke solution with class distinction for LLP, which risks suboptimal performance in the face of poor dataset semantics and a black-box bagging scenario.

\subsection{Contrastive Learning on Tabular Data.}
Contrastive learning has become a potent Self-supervised learning paradigm in computer vision and natural language processing \cite{zhai2019s4l, tung2017self, devlin2018bert, jing2020self}, leveraging positive and negative pairs to optimize embedding spaces. Typical techniques like SimCLR \cite{chen2020simple} utilize data augmentations to create multiple views from single instances, employing InfoNCE loss \cite{gutmann2010noise, oord2018representation} for the learning process. This approach has also been adapted for supervised learning, shifting focus from homologous vs. non-homologous to class distinction \cite{khosla2020supervised}. However, applying contrastive learning to tabular data presents unique challenges due to low-quality data, lack of spatial correlations, and heterogeneity \cite{grinsztajn2022tree, borisov2022deep}. Generating label-invariant views or augmentations in tabular datasets requires preserving essential properties while introducing variation, a non-trivial task barely addressed through techniques like feature masking and random manipulation \cite{yao2021self, yoon2020vime}. VIME \cite{yoon2020vime} capitalizes on corruption techniques tailored for tabular data and introduces a consistency loss. This corruption concept has later extended to contrastive learning by SCARF \cite{bahri2021scarf} based on empirical distributions. SAINT \cite{somepalli2021saint} further combines contrastive learning with denoising reconstruction, integrating feature mixup augmentation \cite{zhang2017mixup, yun2019cutmix}. However, the mixed nature of tabular data, along with issues like significant predictive power held by individual features, makes any form of data manipulation or augmentation a sensitive operation \cite{shavitt2018regularization}. Recent works propose to subset the raw features \cite{ucar2021subtab} to tackle these challenges, but generating strong and intact augmentations for Contrastive learning, even in fully supervised scenarios, remains a challenge.

\section{Methodology}
\subsection{Motivational Study and Overview}
\subsubsection{Inherent LLP Difficulty}
\label{ssec:methodology_difficulty}
As introduced in Sec.\ref{sec:introduction}, a pivotal LLP difficulty is the inherent ambiguity when multiple instance label combinations within a bag yield identical label proportions \cite{la2022learning}, leading to a misalignment between bag-level training objectives and instance-level predictions. This discrepancy hinders the model's ability to classify individual instances from aggregated data.

The concavity of the logarithm function implies, according to Jensen's Inequality \cite{jensen1906fonctions}, that for a concave function $f$ and a random variable $X$, $f(\mathbb{E}[X]) \geq \mathbb{E}[f(X)]$. Applying this principle to the LLP loss highlights a critical insight. Given the instance-level classification formulated as the cross-entropy loss:
\begin{equation}
	\mathcal{L}_{\text{Ins}}(\mathbf{\hat{y}}, \mathbf{y}) = -\frac{1}{m} \sum_{i=1}^{m} \sum_{c=1}^{C} y_{i}^c \log \hat{y}_{i}^c
\end{equation}
where $y_{i}^c$ is the true label of instance $i$ for class $c$, and $\hat{y}_{i}^c=f_\theta(x_{i})^c$ is the predicted probability of instance $i$ belonging to class $c$. We observe that $\mathcal{L}_{\text{LLP}}$'s logarithm of the average prediction $\log(\mathbb{E}[X])$ is greater than or equal to $\mathcal{L}_{\text{Ins}}$'s average of the logarithms of individual predictions $\mathbb{E}(\log[X])$. This mathematical exposition underscores a fundamental gradient suboptimality \cite{khosla2020supervised} of relying solely on bag-level loss functions, such as LLP loss \cite{ardehaly2017co, dulac2019deep}, for optimizing instance-level predictions. While such an approach may align the model's average predictions with the true label proportions across bags, it does not inherently guarantee the precision of predictions for each instance within those bags. Ensuring models achieve both aggregate accuracy and instance-level precision requires strategies that directly address the optimization of individual instance predictions within the LLP framework.

\subsubsection{Tabular-based LLP Problem Formulation}
We defined the tabular LLP Problem addressed by TabLLP-BDC as:
\begin{Definition}[TabLLP-BDC's tabular-based LLP Problem]
	\label{def:1}
	Given an instance-level classifier in deep tabular learning that consists of an encoder $f_{\theta}$, projection head $g_{\delta}$, prediction head $h_{\omega}$, a tabular LLP dataset $D_{LLP} = \left\{ (B_k, \mathbf{\bar{p}_k}) \right\}_{k=1}^{K}$, and loss weights $\alpha$, $\beta$, $\lambda$, $\gamma$, the definition of TabLLP-BDC is formally formulated as:
	\begin{align*}
		\theta_{init},\! \delta^* &\!=\!\underset{\theta, \delta}{\text{arg min}} \left[ \alpha \mathcal{L}_{Bag}(g_{\delta}(f_{\theta}(B)), \mathbf{\bar{p}}) \! + \beta \!\mathcal{L}_{Self}(g_{\delta}(f_{\theta}(B))) \right], \\
		\theta^*, \omega^* &= \underset{\theta_{init}, \omega}{\text{arg min}} \left[ \lambda  \mathcal{L}_{Diff}(f_{\theta}(B), \mathbf{\bar{p}}) + \gamma  \mathcal{L}_{LLP}(h_{\omega}(f_{\theta}(B)), \mathbf{\bar{p}}) \right],
	\end{align*}
	where $\mathcal{L}_{Bag}(\cdot, \cdot)$ denotes our Bag Contrastive pretraining loss, $\mathcal{L}_{Self}(\cdot)$ denotes the classical Self-contrastive + Denoising Reconstruction pretraining loss, $\mathcal{L}_{Diff}(\cdot, \cdot)$ denotes our instance-level Difference Contrastive loss, and $\mathcal{L}_{LLP}(\cdot, \cdot)$ denotes the classical bag-level LLP loss.
\end{Definition}
Compared to classical LLPs, the novelty of TabLLP-BDC includes:
\begin{itemize}[leftmargin=*,nosep]
	\item An instance-level, \textit{class-aware} fine-tuning strategy $\mathcal{L}_{Diff}(\cdot, \cdot)$ on the encoded representation $f_{\theta}(B)$ to bridge the gap between bag-level training ($\mathcal{L}_{LLP}(\cdot, \cdot)$) and instance-level prediction.
	\item A bag-level pretraining strategy $\mathcal{L}_{Bag}(\cdot, \cdot)$ on top of the projection $g_{\delta}(f_{\theta}(B))$ to ensure intrinsic tabular feature correlations captured by $\mathcal{L}_{Self}(\cdot)$ comply with label proportion distribution.
\end{itemize}

\begin{figure}[!t]
	\centering
	\includegraphics[width=\linewidth]{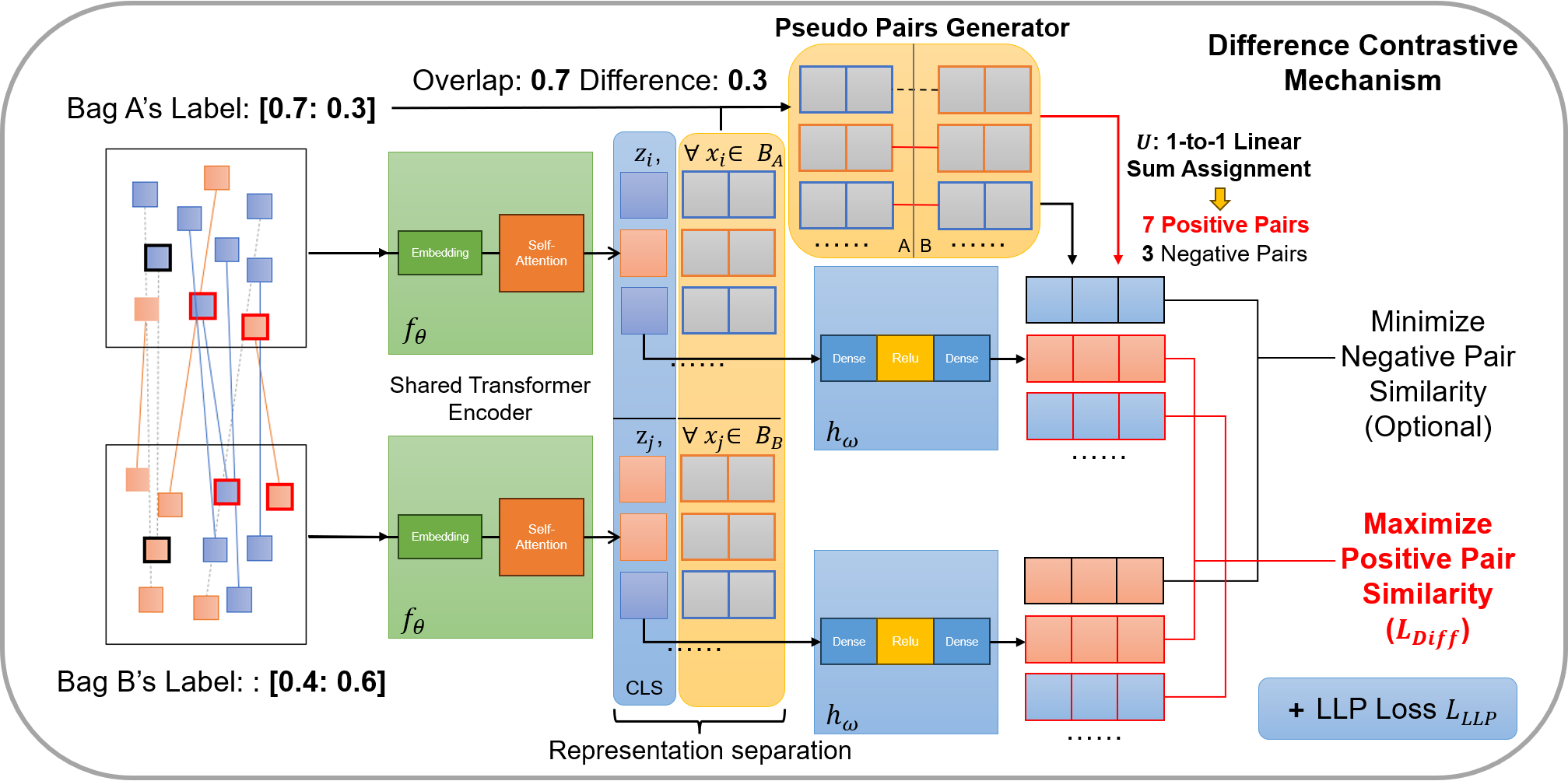}
	\caption{Overview of TabLLP-BDC's Difference Contrastive Mechanism. Consider the optimal mapping between two instance bags, colored lines represent potential positive samples, and dashed lines represent negative samples. The figure shows the forward process of two positive sample pairs marked in red and one negative sample pair marked in black.}
	\label{fig:diff_arch}
	\vspace{-10px}
\end{figure}

Algo.~\ref{alg:tabllpbdc} presents the procedure of two-stage training. Both stages are favorable for tabular data by relying on bag comparison, thus achieving augmentation-free, and leveraging label proportion to establish contrastive signals, thus achieving \textit{class-aware}. See Appx.~\ref{sec:backbone}, Fig.~\ref{fig:diff_arch}, and \ref{fig:bag_arch} for the detailed model architecture.

\subsection{Difference Contrastive Finetuning}
\label{sec:diff_contrastive}
Based on the theoretical insights into the inherent LLP difficulty (Sec.~\ref{ssec:methodology_difficulty}), we propose a Difference Contrastive mechanism to augment the missing instance-level supervision that arises when relying solely on bag-level finetuning loss $\mathcal{L}_{LLP}(\cdot, \cdot)$. This method is anchored on an inherent property of label proportion observed during multi-bag comparisons. Given an optimal encoder $f_{\theta}^*$ and two bags of instances with label proportion $\mathbf{\bar{p}}_1$ and $\mathbf{\bar{p}}_2$, there exists an optimal 1-to-1 mapping assignment between the instance representations from different bags, forming $n_{\text{pos}} = m \sum_{c=0}^{C} \min(\bar{p}_{1}^c, \bar{p}_{2}^c)$ positive pairs of the same label and $n_{\text{neg}} = m - n_{\text{pos}}$ negative pairs of different labels (depicted in Fig.~\ref{fig:diff_arch}).

Formally, for two bags of samples, each of size $n$, and a similarity matrix $S$ where $S_{ij}$ represents the cosine similarity between the $i$-th sample of the first bag and the $j$-th sample of the second, the optimal assignment aims to determine the positive pairs set $P$ and the negative pairs set $Q$ to maximize the given formula:
\begin{equation*}
	\begin{aligned}
		& O(S, n_{\text{pos}}) = \max_{P, Q} \left( \sum_{(i, j) \in P} S_{ij} + \sum_{(i, j) \in Q} -S_{ij} \right) \\
		&\text{s.t. } 
		%		\begin{aligned}[t]
			%			&|P| = n_{\text{pos}}, \\
			%			&|Q| = s - n_{\text{pos}}, \\
			%			&U = P \cup Q \text{ is a 1-to-1} \\
			%			&\text{linear assignment.}
			|P| = n_{\text{pos}},
			|Q| = m - n_{\text{pos}}, \\
			&\ \ \ \ \ \ U = P \cup Q \text{ is a 1-to-1}
			\text{ linear assignment.}
		\end{aligned}
\end{equation*}
The resulting positive and negative pairs can then be used as pseudo-labels to train the model with the Cosine Embedding loss to encourage positive pairs to have similar representation and vice versa. Unfortunately, this problem is inherently combinatorial, and cannot be solved by a computationally efficient algorithm. A notable exception arises when two bags have identical label proportions. In such a case, $n_{\text{pos}}=m$ and no negative pairs exist, the formula can therefore be simplified as:
\begin{equation*}
		\begin{aligned}
			& O(S) = \max_{U} \sum_{(i, j) \in U} S_{ij} \\
			& \text{s.t. } U \text{ is a 1-to-1} \text{ linear assignment.}
		\end{aligned}
\end{equation*}
This is known as the Linear Sum Assignment problem and can be solved by the Hungarian Algorithm \cite{crouse2016implementing} in polynomial time. By removing the explicit consideration of negative pairs, this method yields a good approximation of our combinatorial problem. 

Therefore, given the positive pairs generated by selecting the Top-$n_{\text{pos}}$ similarity 1-to-1 pairs after adopting the Hungarian Algorithm, we propose the Difference Contrastive Loss as follows:
\begin{equation}
		\mathcal{L}_{Diff}(\mathbf{Z}, P) = \frac{1}{|P|} \sum_{(i, j) \in P} -\log \frac{\exp(\text{cos}(\mathbf{z}_i, \mathbf{z}_j) / \tau)}{\sum_{k \in \text{another bag}} \exp(\text{cos}(\mathbf{z}_i, \mathbf{z}_k) / \tau)}
\end{equation}
where $\mathbf{z_i} \in \mathbf{Z}$ is the instance representation, $(i, j)$ is a positive pair from $P$, $\text{cos}$ is the cosine similarity, and $k$ is any sample from the other bag including $j$. This Difference Contrastive Loss utilizes the pseudo-pairs generated based on the difference in two bags' label proportions, serving as an intermediary between the classical Self-contrastive loss \cite{chen2020simple} and the Supervised Contrastive loss \cite{khosla2020supervised}. In contrast to the NT-Xent loss, which is typically employed as the Self-contrastive loss, our approach shifts the contrastive objective from determining if representations originate from the same sample to discerning if they belong to the same class, which is, therefore, \textit{class-aware}. This adjustment not only integrates instance-level supervision with \textit{class-aware} guidance into the LLP task but also obviates the need for label-invariant augmentation to produce varied views of instances. All instances from the opposite bag of anchor instances $\mathbf{z}_i$ are treated as negative pairs because of the inherent 1-to-1 mapping constraint and the approximation assumption.
	
The overall finetuning loss is defined in a multi-task fashion as a ramp-down LLP Loss and a ramp-up Difference Contrastive Loss: 
\begin{equation}
	\begin{aligned}
		\mathcal{L} &= \lambda (t) \mathcal{L}_{Diff}(\mathbf{Z}, P) + \gamma (t) \mathcal{L}_{LLP}(\mathbf{\hat{p}}, \mathbf{\bar{p}}) \\
		&= \lambda (t) \frac{1}{|P|} \sum_{(i, j) \in P} -\log \frac{\exp(\text{cos}(\mathbf{z}_i, \mathbf{z}_j) / \tau)}{\sum_{k \in \text{another bag}} \exp(\text{cos}(\mathbf{z}_i, \mathbf{z}_k) / \tau)} \\
		&+ \frac{\gamma (t)}{2} [ \sum_{c=1}^{C} \mathbf{\bar{p}_1^c} \log  \frac{\mathbf{\bar{p}_1^c}}{\mathbf{\hat{p}_1^c}} + \sum_{c=1}^{C} \mathbf{\bar{p}_2^c} \log  \frac{\mathbf{\bar{p}_2^c}}{\mathbf{\hat{p}_2^c}} ]
	\end{aligned}
\end{equation}
where $\lambda (t) = \exp\left(-5 \times \left(1 - \frac{t}{T}\right)^2\right)$ and $\gamma (t) = 1 - \lambda (t)$ are the exponential ramp-up and ramp-down weight at epoch $t$ over $T$. $\mathcal{L}_{LLP}(\cdot, \cdot)$ serves as an initiator for learning an optimal classifier $f_\theta^*$, aiming at generating pseudo-positive pairs more accurately. 

\begin{table*}[!t]
	\caption{Fine-grained Experimental Results.}
	\label{tab:fine_grained_result}
	\vspace{-5px}
	\small
	\begin{tabular}{l@{\hspace{8pt}}c@{\hspace{8pt}}c@{\hspace{8pt}}c@{\hspace{8pt}}c@{\hspace{8pt}}c@{\hspace{8pt}}c@{\hspace{8pt}}c@{\hspace{8pt}}|@{\hspace{8pt}}c@{\hspace{8pt}}c}
		\toprule
		& \multicolumn{7}{c}{AUC (\%)} & \multicolumn{2}{c}{Accuracy (\%)} \\
		\cmidrule(lr){2-10}
		& \textbf{AD} & \textbf{BA} & \textbf{CA} & \textbf{CR} & \textbf{EL} & \textbf{RO} & \textbf{ML} & \textbf{Private A} & \textbf{Private B} \\
		\midrule
		Supervised & $91.37 \pm 0.20$ & $94.01 \pm 0.18$ & $95.87 \pm 0.08$ & $82.36 \pm 0.07$ & $90.73 \pm 0.15$ & $89.48 \pm 0.05$ & $86.02 \pm 0.1$ & $60.02 \pm 1.54$ & $86.77 \pm 0.40$ \\
		\midrule
		DLLP & $87.61 \pm 0.21$ & $77.18 \pm 1.07$ & $78.75 \pm 1.50$ & $75.07 \pm 0.58$ & $74.33 \pm 0.71$ & $79.10 \pm 0.19$ & $73.75 \pm 0.81$ & $50.39 \pm 0.18$ & $68.86 \pm 2.63$ \\
		LLP-GAN & $83.60 \pm 1.25$ & $72.06 \pm 2.45$ & $83.05 \pm 0.55$ & $75.90 \pm 1.07$ & $56.75 \pm 3.74$ & $74.44 \pm 0.40$ & $74.31 \pm 1.52$ & $50.14 \pm 0.15$ & $66.77 \pm 4.14$ \\
		LLP-VAT & $86.76 \pm 0.25$ & $81.19 \pm 2.03$ & $79.99 \pm 2.17$ & $74.93 \pm 0.45$ & $74.34 \pm 0.76$ & $79.09 \pm 0.22$ & $71.12 \pm 1.20$ & $49.47 \pm 0.39$ & $73.62 \pm 0.93$ \\
		SelfCLR-LLP & $87.44 \pm 0.22$ & $79.73 \pm 1.51$ & $81.34 \pm 2.11$ & $75.43 \pm 0.34$ & $74.44 \pm 0.81$ & $78.97 \pm 0.30$ & $74.36 \pm 1.31$ & $50.28 \pm 0.24$ & $68.99 \pm 2.88$ \\
		\midrule
		TabLLP-DEC & $62.62 \pm 3.35$ & $70.24 \pm 1.53$ & $81.61 \pm 0.32$ & $76.27 \pm 0.42$ & $73.71 \pm 2.47$ & $72.68 \pm 0.44$ & $55.34 \pm 0.50$ & $49.77 \pm 0.04$ & $55.70 \pm 3.39$ \\
		TabLLP-PSE & $87.64 \pm 0.29$ & $80.26 \pm 1.67$ & $79.73 \pm 1.85$ & $77.53 \pm 0.72$ & $75.42 \pm 1.91$ & $79.06 \pm 0.10$ & $74.18 \pm 0.36$ & $50.48 \pm 0.13$ & $70.37 \pm 1.17$ \\
		TabLLP-SELF & $86.66 \pm 0.38$ & $79.38 \pm 0.90$ & $82.55 \pm 0.74$ & $77.04 \pm 0.97$ & $74.87 \pm 0.78$ & $79.19 \pm 0.28$ & $74.10 \pm 0.79$ & $50.41 \pm 0.28$ & $66.60 \pm 1.64$ \\
		\textbf{TabLLP-BDC} & $ \mathbf{87.81 \pm 0.18} $ & $ \mathbf{84.95 \pm 1.88} $ & $ \mathbf{84.85 \pm 0.93} $ & $ \mathbf{79.29 \pm 0.39} $ & $ \mathbf{77.53 \pm 1.48} $ & $ \mathbf{79.67 \pm 0.25} $ & $ \mathbf{75.06 \pm 0.58} $ & $ \mathbf{51.39 \pm 0.16} $ & $ \mathbf{74.03 \pm 1.54} $ \\
		\bottomrule
	\end{tabular}
	\vspace{-5px}
\end{table*}

\subsection{Bag Contrastive Pretraining}
\label{sec:bag_contrastive}
Learning from Label Proportion (LLP), like other subareas of WSL, holds significant potential for pretraining. By capturing feature correlations before being supervised by ambiguous weak labels, the model is less likely to converge to suboptimal solutions \cite{tsai2022learning}. Despite its potential, the exploration of pretraining specifically for tabular-based LLP framework remains untouched, likely due to the dominant emphasis on image, where many universal pretraining models exist \cite{deng2009imagenet}. In order to fill this gap, we first adopted a proven unsupervised pre-training pipeline in the field of deep tabular learning, termed this approach ``Self-contrastive pretraining.'' As detailed in the Appx.~\ref{sec:pretraining_pool}, this instance-level method harmonizes MixUp \cite{zhang2017mixup} and CutMix \cite{yun2019cutmix} augmentations with Self-contrastive and denoising reconstruction to enhance pretraining efficacy \cite{somepalli2021saint}.

However, empirical studies in Tab.~\ref{tab:pretraining} reveal that such a pretraining approach may occasionally underperform compared to scenarios devoid of pretraining within the LLP context, highlighting its unpredictable effectiveness. To mitigate this discrepancy between instance-level pretraining and bag-level learning, we propose the Bag Contrastive pretraining task. This innovation ensures that the Self-contrastive pretraining process \cite{somepalli2021saint, bahri2021scarf} aligns with the variations in label proportions at the bag level, thereby nurturing suitable sample representations for subsequent fine-tuning tasks. Drawing inspiration from the attention mechanism—particularly intersample attention \cite{vaswani2017attention, somepalli2021saint}—alongside classical MIL aggregation methods \cite{ilse2018attention}, we advocate a bag aggregator to compute the weighted sum of instance embeddings within a bag, where weights are determined by intersample similarity through a query network:
% \begin{equation}
% 	\alpha_i = \frac{\exp\left(\mathbf{W} \mathbf{z}_i \cdot \mathbf{z}_i^T\right)}{\sum_{j=1}^{m} \exp\left(\mathbf{W} \mathbf{z}_j \cdot \mathbf{z}_j^T\right)}, \ \ \mathbf{b} = \sum_{i=1}^{m} \alpha_i \mathbf{z}_i,
% \end{equation}
\begin{equation}
	\mathbf{b} = \text{softmax}\left(\mathbf{Z} \mathbf{W} \mathbf{Z}^T\right) \mathbf{Z}
\end{equation}
where $\mathbf{Z}$ is the matrix of instance representations in a bag, $\mathbf{W}$ denotes the weight matrix of the query network,
% $\alpha_i$ signifies the attention score of the $i$-th instance influencing its contribution to the bag representation, 
and $\mathbf{b}$ is the aggregated bag representation. This aggregation adeptly computes the bag representation \cite{ilse2018attention}, and the contribution of each sample to the bag representation is determined by its relation with other samples within the same bag, implicitly reflecting the bag label proportion. To instill reliable bag-level labels, we design the Bag Contrastive task along the principles of deep metric learning \cite{kaya2019deep}:
\begin{align}
	\mathcal{L}_{Bag}(\mathbf{b}, \mathbf{\bar{p}}) &= (1 - \text{mPIoU}(\mathbf{\bar{p}}_1, \mathbf{\bar{p}}_2)) \cdot \max(0, \cos(\mathbf{b}_1, \mathbf{b}_2) - \text{margin}) \nonumber \\
	&\quad + \text{mPIoU}(\mathbf{\bar{p}}_1, \mathbf{\bar{p}}_2) \cdot (1 - \cos(\mathbf{b}_1, \mathbf{b}_2)).
\end{align}
Here, $\mathbf{\bar{p}}_1$ is the label proportion of the first bag, $\mathbf{b}_2$ is the bag representation of the second bag, $\text{mPIoU}$ symbolizes our proposed mean proportion intersection over union metric between two bags (details in Sec.~\ref{sec:evaluation_methods}), while $\text{cos}$ signifies the cosine similarity between the two bag representations. Alternative metrics like L1 can substitute $\text{mPIoU}$ without notable performance deviations, and the efficacy of our proposed bag-level metric is explored in the Appx.~\ref{sec:bag_metrics}. Collectively, the Bag Contrastive loss encourages that the model's representations are consistent with the label proportions, guiding the pretraining toward more reliable embeddings.

\section{Experiments}
\subsection{Experimental Settings}
\subsubsection{Datasets, preprocessing, and bagging.}
Our major study utilizes seven datasets: seven public tabular datasets sourced from OpenML\footnote{\url{https://www.openml.org/}} and a private dataset from the gaming industry. These public datasets, renowned in literature as classical binary classification datasets, each comprise at least 15,000 instances, aligning with the large-scale and black-box scenario of LLP. The private dataset, encompassing in-game behaviors of over 300,000 players, aims to deduce multi-class gamer demographics. Labels for this dataset are exclusively available in a bag label proportion format from an advertising platform, with a constraint of a minimum of 250 distinct instances per bag. We employ the same preprocessing standards as related works \cite{grinsztajn2022tree, somepalli2021saint, bahri2021scarf} and advocate an ordered bagging strategy for scalability and label diversity. Details of dataset statistics (Tab.~\ref{tab:dataset_stats}), preprocessing techniques, and comparison of bagging strategies (Fig.~\ref{fig:bagging_distribution} and Tab.~\ref{tab:bagging_comparison}) are discussed in the Appx.~\ref{sec:datasets_prep} and ~\ref{sec:bagging}. 

\subsubsection{Baselines}
Given the nascent exploration of LLP in tabular data, we adapt several leading LLP image data baselines and introduce intuitive model ideas for pure tabular scenarios. Notably, direct adaptations of some image data baselines may not be feasible due to tabular data's distinct characteristics, so our selection prioritizes methods that are easily transferable. The list of comparing baselines includes: DLLP \cite{ardehaly2017co} (fundamental LLP loss), LLP-GAN \cite{liu2022llp} (superior GAN-based model in the image domain), LLP-VAT \cite{tsai2020learning} (classical consistency regularization work in the image domain), SelfCLR-LLP \cite{nandy2022domain} (the only work addressing deep LLP in the tabular domain), TabLLP-DEC (intuitive adaptation of classical Deep Clustering model \cite{xie2016unsupervised} with LLP constrain), TabLLP-PSE (intuitive pseudo-labeling with constrained KMeans, similar to \cite{liu2021two}), and TabLLP-SELF (intuitive contrastive variant inspired from SelfCLR-LLP \cite{nandy2022domain} but with typical augmentation technique for tabular data \cite{somepalli2021saint, yoon2020vime, yun2019cutmix, zhang2017mixup}). The details of necessary modification and reasons for selection are discussed in the Appx.~\ref{sec:baselines}.

\subsubsection{Evaluation Methods}
\label{sec:evaluation_methods}
Consistent with prevalent practices \cite{ardehaly2017co, liu2022llp, tsai2020learning, nandy2022domain}, our primary evaluation metrics are the AUC score for binary classification and Accuracy for multi-class classification. Recognizing the inherent difficulty of LLP evaluations without instance-level labels, we employ the L1 as an alternative metric \cite{tsai2020learning}. Additionally, we propose the mean Proportional-Intersection-Over-Union (mPIoU) metric, inspired by semantic segmentation's mIoU metric. Given $\hat{\mathbf{p}}^c$ as the predicted label proportion for class $c$, $\bar{\mathbf{p}}^c$ as the true label proportion for class $c$, and $C$ as the number of classes for which the union is non-zero, the mPIoU is defined as:
$
\text{mPIoU} = \frac{1}{C} \sum_{c=1}^{C} \frac{\min(\hat{\mathbf{p}}^c, \bar{\mathbf{p}}^c)}{\max(\hat{\mathbf{p}}^c, \bar{\mathbf{p}}^c)}
$.
This metric evaluates the congruence between predicted and actual label proportions across classes. The metric's efficacy is further discussed in the Tab.~\ref{tab:metrics} and Appx.~\ref{sec:bag_metrics}.

\subsubsection{Model Implementation and Training Setting}
The SAINT-s model \cite{somepalli2021saint} is our default encoder for tabular data. We undergo 50 epochs in pretraining and 300 in finetuning, with a 20-epoch early stopping based on validation scores. Hyperparameters are optimized using Optuna\footnote{\url{https://optuna.org/}} across 50 trials for each dataset and model, and we repeat 10 trials per experiment with varied seeds. Further details and component analysis are discussed in the Appx.~\ref{sec:architecture_training}.

\subsection{Experimental Results}
Our comprehensive experimental evaluation demonstrates the superior performance of TabLLP-BDC against a wide range of LLP baselines across various validation set granularities and bag sizes. For consistency and relevance to practical applications, experiments default to a bag size of 256. Detailed insights into the components' contributions are further explored in the Sec.~\ref{sec:overall_ablation} and Appx.~\ref{sec:ablation}.

\subsubsection{Performance Evaluation with Fine-grained Validation}
This section delves into a common LLP setup featuring accessible fine-grained validation, aligning with evaluations in prior works \cite{ardehaly2017co, liu2022llp, tsai2020learning, nandy2022domain, liu2021two}. Tab.~\ref{tab:fine_grained_result} presents AUC scores for binary datasets and accuracy for multi-class datasets, where TabLLP-BDC exhibits consistent superiority over other LLP classifiers. This advantage is attributed to our novel approach combining Bag Contrastive pretraining and Difference Contrastive fine-tuning, which fosters \textit{class-aware} and instance-specific learning. As discussed in Sec.\ref{ssec:experiment_class_awareness}, unlike methods such as SelfCLR-LLP \cite{nandy2022domain} and TabLLP-SELF, which lack \textit{class-awareness}, TabLLP-BDC explicitly ensures instances of the same class are represented similarly. 
% In Tab.~\ref{tab:movie}, our experiment on the item recommendation task that SelfCLR-LLP focuses on can more intuitively demonstrate the superiority of our model in this regard. 

\subsubsection{Performance Evaluation with Coarse-grained Validation}
Addressing a more challenging scenario devoid of fine-grained validation, often restricted by privacy considerations \cite{GoogleAdsDataHub2023}, this practical analysis validates models' performance based on bag label proportions, implementing early stopping based on L1 scores or our mPIoU metric. Our results, displayed in Tab.~\ref{tab:coarse_grained_result} (Appx.\ref{sec:coarse_grained}), reiterate TabLLP-BDC's lead in performance, albeit with some variations in mPIoU or L1 scores. These fluctuations underscore the complex nature of tabular data, characterized by limited semantics and prevalent noise \cite{grinsztajn2022tree, borisov2022deep}. The imperfect correlation between the best instance-level predictions and label proportions, as shown in Tab.~\ref{tab:coarse_grained_saint} (Appx.\ref{sec:coarse_grained}), validates our hypothesis that solely relying on bag-level loss is insufficient for accurate instance-level prediction in tabular data.

\subsubsection{Verification of Class Awareness}
\label{ssec:experiment_class_awareness}
We assess class awareness in the representation space, introducing the Class Awareness Score (CAS). This metric, reflecting the ratio of standardized intra-class to the sum of standardized intra-class and inter-class cosine similarities, gauges the discriminative capacity of representations. TabLLP-BDC, as depicted in Tab.~\ref{tab:similarity}, exhibits superior discriminative capabilities over the three most potent baselines lacking class awareness across five medium-sized datasets. This underscores the value of \textit{class-aware} supervision in generating distinct and robust class representations. Notably, although DLLP also achieves good CAS, its intra-class and inter-class similarities are very low, which reflects the inherent ambiguity of label proportion matching.

\begin{table}[!t]
	\caption{The Class Awareness Score (CAS) is calculated based on the standardized intra-class and inter-class similarity in the representation space, with higher scores indicating higher \textit{class-awareness}. The highest score for each class is bolded, while the second highest is underlined.}
	\label{tab:similarity}
	\vspace{-5px}
	\begin{tabular}{lccccc}
		\toprule
		& \multicolumn{5}{c}{Class Awareness Score (CAS)} \\
		\cmidrule(lr){2-6}
		\textbf{Model} & \textbf{AD} & \textbf{BA} & \textbf{CA} & \textbf{CR} & \textbf{EL} \\
		\midrule
		DLLP & \textbf{0.8223} & \uline{0.5208} & 0.5192 & \uline{0.8251} & 0.5008 \\
		SelfCLR-LLP & 0.5461 & 0.5167 & \uline{0.5446} & 0.6090 & 0.5005 \\
		TabLLP-SELF & 0.5080 & 0.5013 & 0.5160 & 0.5719 & \uline{0.5103} \\
		\textbf{TabLLP-BDC} & \uline{0.6176} & \textbf{0.6329} & \textbf{0.5802} & \textbf{0.9514} & \textbf{0.5104} \\
		\bottomrule
	\end{tabular}
	\vspace{-15px}
\end{table}

\subsubsection{Effectiveness of Pseudo Pair Generator}
We explore the precision of pseudo-positive pair generation in the Difference Contrastive fine-tuning without reliance on granular labels. Tab.~\ref{tab:pairing} reveals that pseudo-positive pairs, derived from label proportion comparisons, achieve notable accuracy. Unlike simple nearest-neighbor pairings, the Linear Sum Assignment method introduces appropriate difficulty to pair generation, occasionally impacting accuracy but fostering more discriminative and class-congruent representations. This strategic trade-off, as empirically validated in Tab.~\ref{tab:generator} (Appx.\ref{sec:downstream_comp}) against the nearest-neighbor pairings, culminates in an uplift in overall model performance.

\begin{table}[!t]
	\caption{The Accuracy of Pseudo-pairs Generator.}
	\label{tab:pairing}
	\vspace{-5px}
	\small
	\begin{tabular}{lccccc}
		\toprule
		& \multicolumn{5}{c}{Accuracy (\%)} \\
		\cmidrule(lr){2-6}
		\textbf{Scenario} & \textbf{AD} & \textbf{BA} & \textbf{CA} & \textbf{CR} & \textbf{EL} \\
		\midrule
		Fine-grained & \textbf{78.1} & \textbf{85.9} & \textbf{63.8} & \textbf{61.7} & \textbf{67.4} \\
		Coarse-grained & 77.0 & 84.9 & 60.4 & 60.2 & 65.8 \\
		\bottomrule
	\end{tabular}
	\vspace{-8px}
\end{table}

\subsubsection{Influence of Bag Size}
\begin{figure}[!t]
	\centering
	\vspace{-5px}
	\begin{minipage}[b]{0.32\linewidth}
		\includegraphics[width=\linewidth]{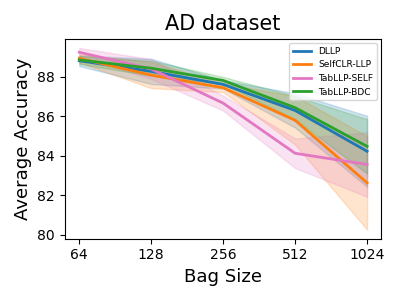}
	\end{minipage}
	\begin{minipage}[b]{0.32\linewidth}
		\includegraphics[width=\linewidth]{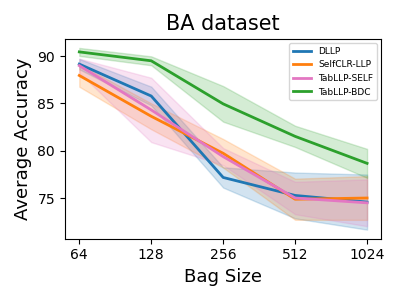}
	\end{minipage}
	\begin{minipage}[b]{0.32\linewidth}
		\includegraphics[width=\linewidth]{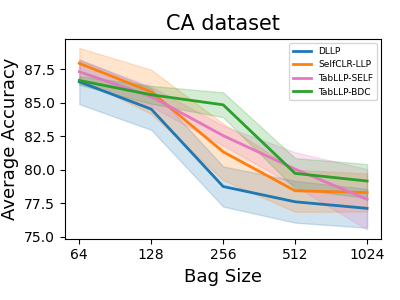}
	\end{minipage}
	\vspace{-1mm}
	\begin{minipage}[b]{0.32\linewidth}
		\includegraphics[width=\linewidth]{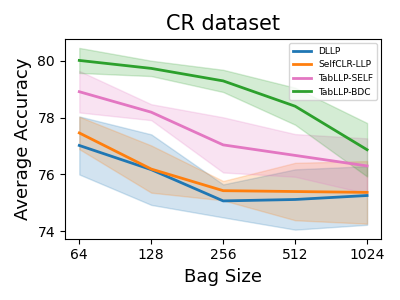}
	\end{minipage}
	\begin{minipage}[b]{0.32\linewidth}
		\includegraphics[width=\linewidth]{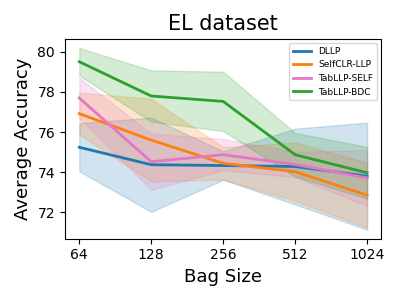}
	\end{minipage}
	\caption{The impact of varying bag sizes on four most potent models across five medium size datasets.}
	\label{fig:bag_sizes}
	\vspace{-10px}
\end{figure}

We examine the impact of varying bag sizes on the LLP framework, prioritizing larger sizes to reflect industry-scale applications: \{64, 128, 256, 512, 1024\}. Fig.~\ref{fig:bag_sizes} illustrates the performance trends and variability of the three most potent baselines within contrastive learning frameworks alongside TabLLP-BDC. Consistent with prior research \cite{ardehaly2017co, liu2022llp, tsai2020learning, nandy2022domain, liu2021two}, prediction accuracy approaches that of supervised learning as bag sizes decrease. Nonetheless, their performance drops markedly with larger bags. Contrary to the sharp performance decline noted in image data LLP studies, tabular-based LLP models, especially our TabLLP-BDC, demonstrate robust adaptability to larger bag sizes, underscoring the promising potential in practical settings.

\begin{figure}[!t]
	\centering
	\vspace{-5px}
	\begin{minipage}[b]{0.32\linewidth}
		\includegraphics[width=\linewidth]{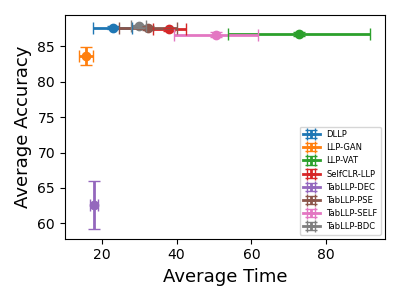}
	\end{minipage}
	\begin{minipage}[b]{0.32\linewidth}
		\includegraphics[width=\linewidth]{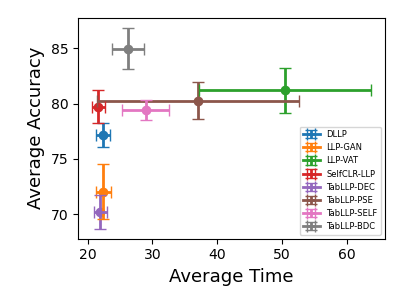}
	\end{minipage}
	\begin{minipage}[b]{0.32\linewidth}
		\includegraphics[width=\linewidth]{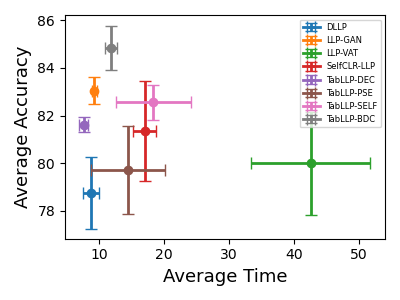}
	\end{minipage}
	\vspace{-1mm}
	\begin{minipage}[b]{0.32\linewidth}
		\includegraphics[width=\linewidth]{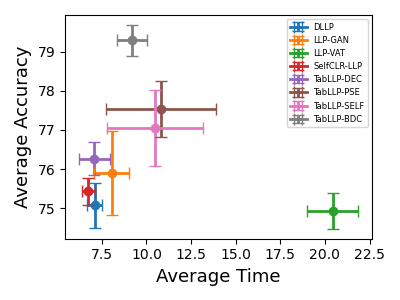}
	\end{minipage}
	\begin{minipage}[b]{0.32\linewidth}
		\includegraphics[width=\linewidth]{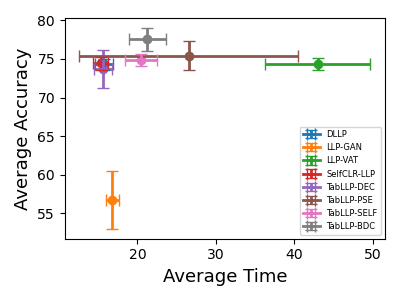}
	\end{minipage}
	\caption{Accuracy-vs-Time diagrams.}
	\label{fig:acc_time_ad}
	\vspace{-15px}
\end{figure}

\subsubsection{Model Robustness}
We conducted a robustness ranking of various models by analyzing their performance standard deviation across multiple datasets, as detailed in Tab.~\ref{tab:model_robustness_ranking}. This ranking system, assigning scores from 1 (most stable) to 8 (least stable) based on standard deviation, serves to quantify each model's stability across diverse scenarios. The overall robustness ranking, derived from averaging these scores, positions TabLLP-BDC as the second most stable model, showcasing its consistent performance. Although DLLP exhibits the highest stability, its model performance across datasets does not match that of TabLLP-BDC. Additionally, the Accuracy-vs-Time diagram for all models, illustrated in Fig.~\ref{fig:acc_time_ad}, demonstrates TabLLP-BDC's balanced trade-off between accuracy and training time, reinforcing its robustness.

\begin{table}[!t]
	\caption{Average robustness ranking for each model.}
	\label{tab:model_robustness_ranking}
	\vspace{-10px}
	\small
	\centering
	\begin{tabular}{l|c}
		\hline
		\textbf{Model} & \textbf{Average Robustness Ranking} \\
		\hline
		DLLP & \textbf{3.375} \\
		LLP-GAN & 6.375 \\
		LLP-VAT & 4.625 \\
		SelfCLR-LLP & 4.5 \\
		TabLLP-DEC & 4.875 \\
		TabLLP-PSE & 4.125 \\
		TabLLP-SELF & 4.5 \\
		\textbf{TabLLP-BDC} & \textbf{3.625} \\
		\hline
	\end{tabular}
	\vspace{-10px}
\end{table}

\subsubsection{Training Time Analysis}
Tab.~\ref{tab:training_time_analysis} presents an evaluation of the training efficiency for various models, highlighting both the average training duration and its variability. This analysis is crucial for gauging each model's computational demands, a key consideration for their deployment in environments with limited resources. For the five medium-sized datasets, TabLLP-BDC achieves Top 2 efficiency in four instances, indicating a remarkable compromise between computational time and performance enhancement. Although the inclusion of a pretraining phase marginally extends the training period, the consistent performance gains across models justify this additional investment. Consequently, TabLLP-BDC emerges as a cost-effective solution for tackling the tabular-based LLP challenge, given its superior accuracy and efficiency.

\begin{table}
	\caption{Average training time and standard deviation.}
	\label{tab:training_time_analysis}
	\vspace{-10px}
	\footnotesize
	\begin{tabular}{l@{\hspace{3pt}}c@{\hspace{3pt}}c@{\hspace{3pt}}c@{\hspace{3pt}}c@{\hspace{3pt}}c}
		\toprule
		& \multicolumn{5}{c}{Avg. (std) Training Time (min)} \\
		\cmidrule(lr){2-6}
		\textbf{Model} & \textbf{AD} & \textbf{BA} & \textbf{CA} & \textbf{CR} & \textbf{EL} \\
		\midrule
		DLLP & $22.86 (5.28)$ & $22.35 (1.11)$ & \uline{$8.76 (1.30)$} & $7.09 (0.44)$ & $15.68 (1.14)$ \\
		LLP-GAN & $22.50 (2.50)$ & $22.45 (1.10)$ & $9.30 (0.34)$ & $8.04 (1.00)$ & $16.80 (0.82)$ \\
		LLP-VAT & $72.80 (19.03)$ & $50.49 (13.24)$ & $42.58 (9.22)$ & $20.43 (1.42)$ & $42.98 (6.73)$ \\
		SelfCLR & $38.06 (4.35)$ & \uline{$21.66 (0.96)$} & $17.01 (1.84)$ & \uline{$6.73 (0.37)$} & \uline{$15.29 (0.90)$} \\
		DEC & \uline{$17.87 (1.16)$} & $21.98 (1.05)$ & \uline{$7.66 (0.67)$} & $7.07 (0.86)$ & $15.65 (1.16)$ \\
		PSE & $32.35 (7.66)$ & $37.06 (15.57)$ & $14.44 (5.69)$ & $10.82 (3.08)$ & $26.55 (13.95)$ \\
		SELF & $50.64 (11.23)$ & $28.97 (3.65)$ & $18.39 (5.73)$ & $10.46 (2.69)$ & $20.44 (2.06)$ \\
		\textbf{BDC} & \uline{$22.23 (2.01)$} & \uline{$20.02 (2.44)$} & $9.19 (0.97)$ & \uline{$6.83 (0.86)$} & \uline{$12.69 (2.34)$} \\
		\midrule
		\textbf{Pretrain} & $7.64 (0.13)$ & $6.20 (0.82)$ & $2.68 (0.15)$ & $2.36 (0.15)$ & $8.57 (0.90)$ \\
		\bottomrule
	\end{tabular}
	\vspace{-13px}
\end{table}

\subsection{Ablation Study}
\label{sec:overall_ablation}
We present a comprehensive ablation study to assess the individual and combined effect of the components within our TabLLP-BDC model in the Appx.~\ref{sec:ablation}. Here is a summary of the main findings:
\begin{itemize}[leftmargin=*,nosep]
	\item \textbf{Ablation Study of Downstream Contrastive Objectives:} The Difference Contrastive objective consistently outperforms other contrastive-based objectives across most scenarios and datasets and achieves solid SOTA when combined with the best pretraining task. (Tab.~\ref{tab:downstream} in Appx.~\ref{sec:downstream_obj})
	\item \textbf{Ablation Study of Downstream Components:} The Linear Sum Assignment method for generating pseudo-positive pairs (Tab.~\ref{tab:generator}), combined with the Supervised InfoNCE-based Contrastive loss (Tab.~\ref{tab:loss}), provides the best balance between performance and efficiency compared to Greedy Pairwise Assignment and Cosine Embedding loss. (Appx.~\ref{sec:downstream_comp})
	\item \textbf{Ablation Study of Pretraining Tasks:} Our pretraining strategy, including the Bag Contrastive task, effectively bridges instance-level pretraining and bag-level finetuning, consistently enhancing the performance of several tabular LLP models. However, combining Bag Contrastive with Difference Contrastive may not always amplify each other's benefits, suggesting potential overlaps. (Tab.~\ref{tab:pretraining} and Appx.~\ref{sec:pretraining_task})
	\item \textbf{Ablation Study of Pretraining Components:} Separating representation by the CLS token is an augmentation-free and more efficient alternative to the typical sample manipulation, e.g., feature mixup, when augmentating tabular data (Tab.~\ref{tab:augmentation}). In terms of aggregation techniques (Tab.~\ref{tab:aggregation}), the similarity-driven weighted sum aggregator generally outperforms the intersample attention-based aggregator both with and without the projector. (Appx.~\ref{sec:pretraining_comp})
	\item \textbf{Evaluation of Bag-Level Metrics in Early Stopped LLP Models:} Both the L1 metric and the proposed mPIoU metric offer valuable insights, but the mPIoU metric provides a more nuanced evaluation, suggesting its potential for a more comprehensive assessment in future research. (Tab.~\ref{tab:metrics} in Appx.~\ref{sec:bag_metrics})
\end{itemize}

\begin{table}
	\caption{Ablation Study of Pretraining Tasks.}
	\label{tab:pretraining}
	\vspace{-10px}
	\begin{tabular}{llccccc}
		\toprule
		& & \multicolumn{5}{c}{AUC (\%)} \\
		\cmidrule(lr){3-7}
		\textbf{Pretrain} & \textbf{Model} & \textbf{AD} & \textbf{BA} & \textbf{CA} & \textbf{CR} & \textbf{EL} \\
		\midrule
		No & DLLP & 87.64 & 77.18 & 78.75 & 75.07 & 74.33 \\
		Self & DLLP & 87.55 & \textbf{84.36} & 83.53 & 75.58 & 75.31 \\
		Bag & DLLP & 87.71 & 81.57 & 83.28 & \textbf{75.60} & 74.82 \\
		Self+Bag & DLLP & \textbf{87.76} & 82.63 & \textbf{84.40} & 75.51 & \textbf{75.64} \\
		\midrule
		No & SelfCLR & 87.44 & 79.73 & 81.34 & 75.43 & 74.44 \\
		Self & SelfCLR & 87.51 & \textbf{83.50} & 84.41 & 75.85 & 75.26 \\
		Bag & SelfCLR & 87.49 & 79.05 & 84.27 & 75.51 & 75.14 \\
		Self+Bag & SelfCLR & \textbf{87.67} & 81.69 & \textbf{84.78} & \textbf{76.13} & \textbf{75.40} \\
		\midrule
		No & SELF & 86.66 & 79.38 & 82.55 & 77.04 & 74.87 \\
		Self & SELF & 86.63 & \textbf{83.16} & 84.04 & 77.09 & 75.22 \\
		Bag & SELF & 86.50 & 77.71 & 82.92 & 77.34 & 75.28 \\
		Self+Bag & SELF & \textbf{86.99} & 82.78 & \textbf{84.19} & \textbf{77.69} & \textbf{75.44} \\
		\midrule
		No & BDC & 87.03 & 81.77 & 80.64 & 79.10 & 75.43 \\
		Self & BDC & 87.79 & \textbf{84.95} & \textbf{84.85} & \textbf{79.29} & 76.24 \\
		Bag & BDC & 87.80 & 84.06 & 83.12 & 78.59 & 75.82 \\
		Self+Bag & BDC & \textbf{87.81} & 84.91 & 83.94 & 79.16 & \textbf{77.53} \\
		\bottomrule
	\end{tabular}
	\vspace{-10px}
\end{table}

\section{Conclusion}
In this study, we addressed the intricate challenges of Learning from Label Proportions (LLP) with a focus on tabular data. Our innovative TabLLP-BDC model introduces a \textit{class-aware} and augmentation-free contrastive learning framework tailored for tabular-based LLP. This approach effectively bridges the gap between bag-level supervision and the desired instance-level predictions, overcoming the unique hurdles of generating \textit{class-aware} and augmentation-free instance-level signals for tabular data. The model's design is underpinned by two key mechanisms: Bag Contrastive pretraining and Difference Contrastive fine-tuning. Together, these elements enable our model to leverage detailed instance-level insights while adhering to holistic bag-level cues. Additionally, our work introduces the mPIoU metric, specifically designed for bag-level validation, enhancing the robustness of our evaluation process and underscoring the practical utility of our model. Empirical validations across various datasets demonstrate TabLLP-BDC's outstanding performance, proving its versatility and efficacy, particularly in user modeling and personalization applications. Looking ahead, this research paves the way for further advancements in pseudo-pair generation techniques based on label proportion comparisons and aims to refine our objectives to better extract and utilize the latent instance-level insights embedded within label proportions.

\newpage

% \begin{acks}
% TBD
% \end{acks}

\bibliographystyle{ACM-Reference-Format}
\bibliography{TabLLP_Draft}

\clearpage
\newpage
\appendix

\section{Methodology}

\subsection{Pseudo Code.}
\label{sec:algorithm}
We present the pseudo-code for our proposed TabLLP-BDC methodology, encapsulated in Algo.~\ref{alg:tabllpbdc}. The algorithm details the complete process of our approach, divided into two distinct phases: Bag Contrastive Pretraining and Difference Contrastive Fine-tuning.

During the Bag Contrastive Pretraining phase, the algorithm iteratively processes pairs of bags from the LLP dataset $D$. For each pair, it computes embeddings for all instances using the encoder $f_{\theta}$ and projects these embeddings via $g_{\delta}$. It then calculates the aggregated bag representations using the weighted sum aggregation mechanism discussed in Sec.~\ref{sec:bag_contrastive}. The pretraining loss comprises the Bag Contrastive loss $\mathcal{L}_{Bag}$, the Self-contrastive loss, and the Denoising Reconstruction loss.

In the Difference Contrastive Fine-tuning phase, the model further refines its understanding of the data. For each pair of bags, embeddings are again computed for all instances. Pseudo-positive and pseudo-negative pairs are generated based on label proportions, facilitating the creation of instance-level supervision signals. The fine-tuning loss is a weighted combination of the Difference Contrastive loss $\mathcal{L}_{Diff}$ and the classical LLP loss $\mathcal{L}_{LLP}$. The exponential ramp-up and ramp-down weights $\lambda (t)$ and $\gamma (t)$ dynamically adjust the emphasis between two losses across fine-tuning epochs.

\begin{algorithm}
	\caption{TabLLP-BDC Methodology}
	\label{alg:tabllpbdc}
	\small
	\DontPrintSemicolon  % Don't print semicolon
	\KwIn{LLP Dataset $D = \{(B_k, \mathbf{\bar{p}_k})\}_{k=1}^{K}$}
	\KwOut{Optimized model parameters $\theta^*, \omega^*$}
	
	\tcc{Phase 1: Bag Contrastive Pretraining}
	\ForEach{pretraining epoch $e$}{
		\ForEach{pair of bags $(B_{k1}, B_{k2})$ from $D$}{
			Compute embeddings $\mathbf{z}_i = f_{\theta}(x_i)$ for all $x_i \in B_{k1} \cup B_{k2}$\;
			Compute projection $\mathbf{j}_i = g_{\delta}(\mathbf{z}_i)$\;
			Compute bag representation $\mathbf{b}_{k1}, \mathbf{b}_{k2}$ using aggregation mechanism\;
			Update $\theta, \delta$ by minimizing $\alpha \mathcal{L}_{Bag}(\{\mathbf{b}_{k1}, \mathbf{b}_{k2}\}, \{\mathbf{\bar{p}}_{k1}, \mathbf{\bar{p}}_{k2}\}) + \beta  \mathcal{L}_{Self}(J)$\;
		}
	}
	
	\tcc{Phase 2: Difference Contrastive Fine-tuning}
	\ForEach{fine-tuning epoch $t$}{
		\ForEach{pair of bags $(B_{k1}, B_{k2})$ from $D$}{
			Compute embeddings $\mathbf{z}_i = f_{\theta}(x_i)$ for all $x_i \in B_{k1} \cup B_{k2}$\;
			Generate pseudo-positive and pseudo-negative pairs $P, Q$\;
			Compute prediction logic $\mathbf{\hat{y}}_i = h_{\omega}(\mathbf{z}_i)$ and then proportion $\mathbf{\hat{p}_k}$\;
			Update $\theta, \omega$ by minimizing $\lambda (t) \mathcal{L}_{Diff}(\mathbf{Z}, P) + \gamma (t) \mathcal{L}_{LLP}(\{\mathbf{\hat{p}}_{k1}, \mathbf{\hat{p}}_{k2}\}, \{\mathbf{\bar{p}}_{k1}, \mathbf{\bar{p}}_{k2}\})$\;
		}
	}
	
	\Return $\theta^*, \omega^*$\;
\end{algorithm}

\subsection{Model Architecture.}
\label{sec:backbone}

Our primary objective is to learn a robust, discriminative, and \textit{class-aware} representation of tabular data in the LLP landscape. To address the challenges inherent in tabular-based LLP, we introduce a novel architecture that leverages the strengths of transformer models \cite{vaswani2017attention} while introducing new components tailored to our LLP paradigm. In our proposed architecture, we utilize a SOTA transformer model called SAINT \cite{somepalli2021saint} in the deep tabular learning domain as the backbone encoder. Building upon SAINT's success in supervised and semi-supervised settings, this design prioritizes both robustness and performance, emphasizing the extraction of high-quality tabular representations. The encoder processes input embeddings using multi-head self-attention and intersample attention, mechanisms that highlight the importance of individual features and capture intricate relationships within or between the data. Mathematically, the SAINT model can be described by its forwarding equations \cite{somepalli2021saint}:
\begin{align*}
	\mathbf{z}^{(1)}_i &= \text{LN}(\text{MSA}(E(x_i))) + E(x_i) \\
	\mathbf{z}^{(2)}_i &= \text{LN}(\text{FF}_1(\mathbf{z}^{(1)}_i)) + \mathbf{z}^{(1)}_i  \\
	\mathbf{z}^{(3)}_i &= \text{LN}(\text{MISA}(\{\mathbf{z}^{(2)}_i\}_{i=1}^{b})) + \mathbf{z}^{(2)}_i \\
	\mathbf{z}_i &= \text{LN}(\text{FF}_2(\mathbf{z}^{(3)}_i)) + \mathbf{z}^{(3)}_i 
\end{align*}
where $\mathbf{z}_i$ is SAINT's contextual representation output corresponding to data point $x_i$, $MSA(\cdot)$ represents multi-head self-attention, $MSA(\cdot)$ represents multi-head intersample attention, $\text{FF}$ denotes feed-forward layers, $\text{LN}$ stands for layer normalization, and $b$ is the batch size. The embedding layer, $E$, is a unique aspect of SAINT, using different embedding functions for different features and missing value embedding for imputation. Let $x_i=[[\text{CLS}], f_i^{\{1\}}, ..., f_i^{\{n\}}]$ be a single data-point with categorical or continuous features, the model exclusively channels the CLS token into an MLP-based prediction head to produce the logic for classification.

\begin{figure}[!t]
	\centering
	\includegraphics[width=\linewidth]{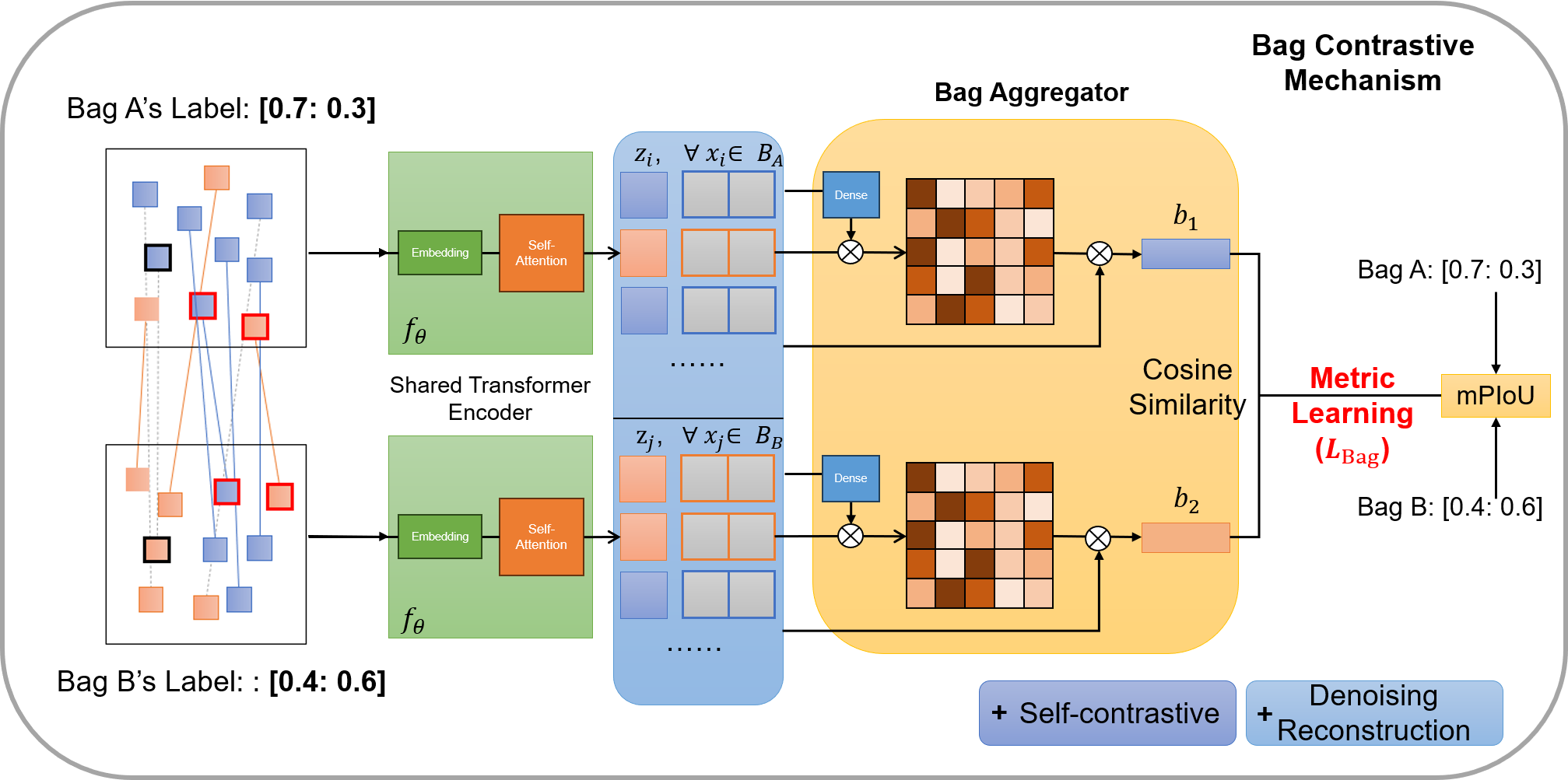}
	\caption{Overview of TabLLP-BDC's Bag Contrastive Mechanism.}
	\label{fig:bag_arch}
	\vspace{-10px}
\end{figure}

To further refine our data representations for LLP during pretraining, we have designed a Bag Aggregator, as illustrated in Fig.~\ref{fig:bag_arch}. This component crafts a bag representation by aggregating instance representations, weighted based on intersample similarity. Its primary function is to emphasize instance-wise similarity in determining the contribution of each instance. Moreover, our model is equipped with a Pseudo-pairs Generator for the Difference Contrastive learning task, as illustrated in Fig.~\ref{fig:diff_arch}. This component leverages linear programming of the linear sum assignment problem to generate pseudo-positive and negative pairs, augmenting the LLP training process with \textit{class-aware} instance-level supervision signals. In summary, our architecture amalgamates the strengths of the SAINT encoder in learning tabular representation with novel components tailored for our two-stage LLP training strategy, ensuring outstanding performance for our task.

\subsection{Pretraining Pool.}
\label{sec:pretraining_pool}

We incorporate two additional losses from SAINT \cite{somepalli2021saint} into our pre-training pool. The first, a Self-contrastive loss, minimizes the distance between latent representations of two different views of the same data point ($\mathbf{z}_i$ and $\mathbf{z}_i'$), while maximizing the distance between distinct data points ($\mathbf{z}_i$ and $\mathbf{z}_j$, $i \neq j$). This employs the InfoNCE loss \cite{gutmann2010noise, oord2018representation}, a metric from metric-learning literature. The second loss is derived from a denoising task aimed at reconstructing the original data sample from its noisy view \cite{somepalli2021saint}. Given a noisy representation $\mathbf{z}_i'$, the goal is to generate a reconstruction $x_i''$ that minimizes the discrepancy between the original data and its reconstruction. The combined pre-training loss $\mathcal{L}_{Self}$ is:
\begin{align*}
	\mathcal{L}_{Self}(\mathbf{Z}) &= \mathcal{L}_{Self-contrastive}(\mathbf{Z}) + \mathcal{L}_{Reconstruct}(\mathbf{Z})	\\
	\mathcal{L}_{Self-contrastive}(\mathbf{Z}) &= -\sum_{i=1}^{m} \log \frac{\exp(\mathbf{z}_i \cdot \mathbf{z}_i' / \tau)}{\sum_{k=1}^{m} \exp(\mathbf{z}_i \cdot \mathbf{z}_k' / \tau)}	\\
	\mathcal{L}_{Reconstruct}(\mathbf{Z}) &= \kappa \sum_{i=1}^{m} \sum_{j=1}^{n} [\mathcal{L}_j(\text{MLP}_j(\mathbf{z}_i'), x_i)]
\end{align*}
Here, $\mathbf{e}_i = E(x_i)$ is the original embedding, $f_{\theta}$ is the encoder, $\mathbf{z}_i=f_{\theta}(\mathbf{e}_i)$, $\mathbf{z}_i'=f_{\theta}(\mathbf{e}_i')$, $\mathbf{z}_i=g_1(\mathbf{z}_i)$, $\mathbf{z}_i'=g_2(\mathbf{z}_i')$, $\kappa $ is a hyper-parameter, and $\tau$ is temperature parameter. $\mathcal{L}$ is cross-entropy loss or mean squared error depending on the $j$th feature being categorical or continuous. The projectors include $n$ single layer $\text{MLP}$ with a ReLU non-linearity.

Following SAINT \cite{somepalli2021saint}, we adopt the combination of MixUp and CutMix as the baseline augmentations to form a challenging yet effective Self-supervision task. Given a data point $x_i$, we compute its original embedding $\mathbf{e}_i = E(x_i)$. The augmented representation, denoted as $\mathbf{e}_i'$, is generated through a combination of MixUp \cite{zhang2017mixup} and CutMix \cite{yun2019cutmix} techniques as follows:
\begin{align*}
	x_i' &= x_i \odot n_m + x_a \odot (1 - n_m)	\\
	\mathbf{e}_i' &= n_u * E(x_i') + (1 - n_u) * E(x_b')
\end{align*}
where $x_a$ and $x_b$ are randomly selected samples from the same batch, $x_b'$ represents the CutMix variant of $x_b$, $n_m$ is a binary mask vector drawn from a Bernoulli distribution with probability $p_{\text{cutmix}}$, and $n_u$ is the MixUp parameter influencing the blend ratio. This effective augmentation strategy stands for the major comparison baseline of our augmentation-free separated representation approach.

\section{Experimental Settings}

\subsection{Datasets and Preprocess.}
\label{sec:datasets_prep}

In total, our study utilizes eight datasets: seven public tabular datasets (five medium sizes, two large sizes) sourced from OpenML\footnote{\url{https://www.openml.org/}} and a private dataset from the gaming industry. The statistics of the datasets are shown in Tab.~\ref{tab:dataset_stats}. We allocate the datasets into an 80\%/10\%/10\% split for training, testing, and validation, respectively. In addition to the fact that the labels of real datasets are inherently in the form of label proportion, we use two different black-box bagging strategies for public datasets to simulate practical application scenarios, as discussed in the following Sec.~\ref{sec:bagging} Bagging Process.

Adhering to the preprocessing standards set by \cite{grinsztajn2022tree, somepalli2021saint, bahri2021scarf}, we employ one-hot encoding for categorical features and Z-normalize numerical features. For public datasets, rather than conventional imputation methods like mean imputation, we leverage the missing value embedding technique from SAINT \cite{somepalli2021saint}. For the private dataset, numerical missing values are imputed with zero due to domain specificity, while missing value embedding is reserved for categorical features.

\begin{table}[!t]
	\small
	\caption{Dataset Statistics.}
	\label{tab:dataset_stats}
	\vspace{-5px}
	\setlength\tabcolsep{2pt}
	\begin{tabular}{lccccc}
		\toprule
		\textbf{Dataset} & \textbf{ID} & \textbf{Size} & \textbf{Numeric} & \textbf{Categorical} & \textbf{Classes} \\
		\midrule
		AD & 1590 & 48842 & 6 & 9 & 2 \\
		BA & 1461 & 45211 & 7 & 10 & 2 \\
		CA & 44090 & 20634 & 8 & 1 & 2 \\
		CR & 44089 & 16714 & 10 & 1 & 2 \\
		EL & 44156 & 38474 & 7 & 2 & 2 \\
		RO & 44161 & 111762 & 29 & 4 & 2 \\
		Movielens-1M & - & 739012 & 3 & 4 & 2 \\
		Private & - & >300000 & 69 & 11 & 2/3 \\
		\bottomrule
	\end{tabular}
	\vspace{-5px}
\end{table}

\begin{figure}[!t]
	\centering
	\begin{minipage}[b]{0.49\linewidth}
		\includegraphics[width=\linewidth]{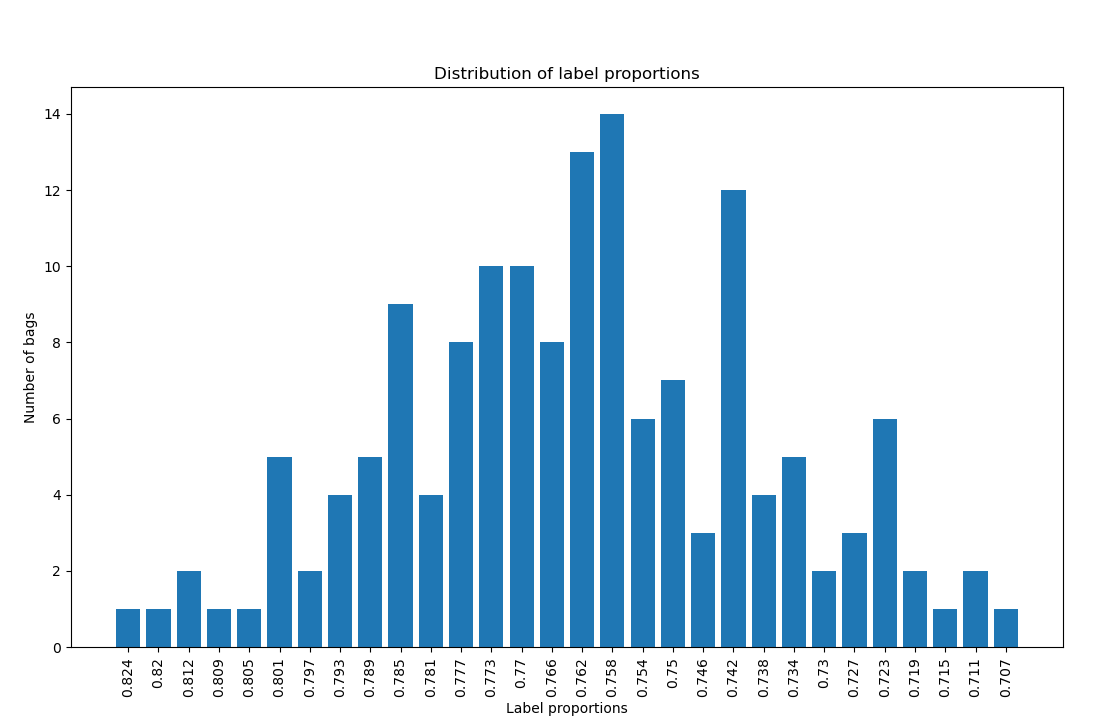}
		\centering
		(a)
	\end{minipage}
	\hfill
	\begin{minipage}[b]{0.49\linewidth}
		\includegraphics[width=\linewidth]{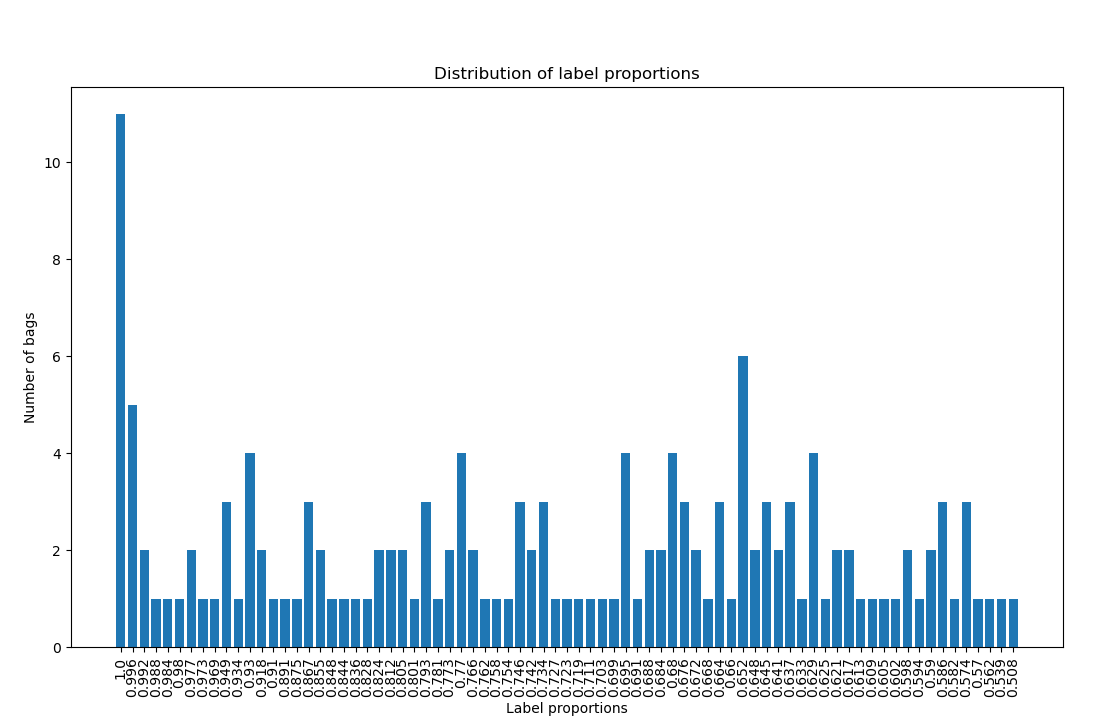}
		\centering
		(b)
	\end{minipage}
	\caption{The bag label proportion distribution after adopting (a) random bagging, and (b) ordered bagging on AD dataset.}
	\label{fig:bagging_distribution}
	\vspace{-10px}
\end{figure}

\begin{table}[!t]
	\footnotesize
	\caption{Analysis of Bagging Strategies across Datasets.}
	\vspace{-5px}
	\label{tab:bagging_comparison}
	\begin{tabular}{ll@{\hspace{8pt}}c@{\hspace{8pt}}c@{\hspace{8pt}}c@{\hspace{8pt}}c@{\hspace{8pt}}c@{\hspace{8pt}}c}
		\toprule
		& & \multicolumn{6}{c}{AUC (\%)} \\
		\cmidrule(lr){3-8}
		\textbf{Model} & \textbf{Strat.} & \textbf{AD} & \textbf{BA} & \textbf{CA} & \textbf{CR} & \textbf{EL} & \textbf{RO} \\
		\midrule
		DLLP & Rand. & \textbf{85.07} & 80.43 & 78.34 & 63.08 & 72.53 & 65.96 \\
		DLLP & \textbf{Ord.} & 87.49 & \textbf{78.27} & \textbf{79.82} & \textbf{75.02} & \textbf{74.90} & \textbf{78.99} \\
		\midrule
		TabLLP-BDC & Rand. & 85.92 & 70.68 & 82.16 & 71.90 & 74.84 & 67.28 \\
		TabLLP-BDC & \textbf{Ord.} & \textbf{87.81} & \textbf{84.95} & \textbf{84.85} & \textbf{79.29} & \textbf{77.53} & \textbf{79.67} \\
		\bottomrule
	\end{tabular}
	\vspace{-10px}
\end{table}

\subsection{Bagging Process}
\label{sec:bagging}

We explore two black-box bagging strategies for the LLP simulation of public datasets: random bagging and ordered bagging. This is because in reality, the label proportion within each bag cannot be validated or explored before bagging and querying the data for response, and no adjustments on bagging can be made post-querying for differential check policy \cite{GoogleAdsDataHub2023}. Random bagging disregards feature correlations and aggregates instances arbitrarily. In contrast, ordered bagging sequences instances based on raw features, grouping similar instances—a rapid variant of clustering for large datasets. Both methods operate under the assumption of no prior domain knowledge. This mirrors the constraints of aggregated reporting \cite{GoogleAdsDataHub2023} and K-anonymity \cite{sweeney2002k}, preventing illegal instance-level data extraction from aggregated datasets. As illustrated in the accompanying Fig.~\ref{fig:bagging_distribution} and Tab.~\ref{tab:bagging_comparison}, ordered bagging consistently surpasses random bagging across various models, attributed to its diverse label proportions. Consequently, our study exclusively employs the ordered bagging strategy.

\subsection{Model Implementation and Training Setting}
\label{sec:architecture_training}
As detailed in Sec.\ref{sec:backbone}, TabLLP-BDC's architecture encompasses an encoder, a CLS token MLP, dual Self-contrastive pretraining projectors, a reconstruction decoder, a bag aggregator, a Bag Contrastive pretraining projector, and a pseudo-pairs generator. Unless otherwise stated, the SAINT model \cite{somepalli2021saint} serves as the encoder for all tabular model variants. Despite SAINT's intersample version's superior performance in supervised and semi-supervised tabular data settings, our findings in Tab.~\ref{tab:saint_attention} indicate no performance enhancement from combining intersample attention and self-attention in LLP. Thus, we utilize SAINT's 6-layer multi-head self-attention variant with recommended hyperparameters\footnote{\url{https://github.com/somepago/saint}}. The CLS token MLP and projectors are 3-layer MLP with ReLU activation, while the reconstruction decoder takes separated MLP for each feature token. We implement the bag aggregator as the weighted sum of instance representation based on intersample cosine similarity, as introduced in Sec.~\ref{sec:bag_contrastive}. The pseudo-pairs generator is achieved by the publicly available implementation\footnote{\url{https://docs.scipy.org/doc/scipy/reference/generated/scipy.optimize.linear\_sum\_assignment.html}} of the modified Jonker-Volgenant algorithm \cite{crouse2016implementing}. But we also studied other possible implementations, summarized in Tab.~\ref{tab:model_components}, in the following ablation studies.

\begin{table}
	%\addtocounter{footnote}{-1}
	%\caption{Configurations of SAINT\protect\footnotemark}
	\small
	\caption{Analysis of Transformer Encoder Configurations.}
	\label{tab:saint_attention}
	\begin{tabular}{l@{\hspace{8pt}}c@{\hspace{8pt}}c@{\hspace{8pt}}c@{\hspace{8pt}}c@{\hspace{8pt}}c@{\hspace{8pt}}c}
		\toprule
		& \multicolumn{6}{c}{AUC (\%)} \\
		\cmidrule(lr){2-7}
		\textbf{Architecture} & \textbf{AD} & \textbf{BA} & \textbf{CA} & \textbf{CR} & \textbf{EL} & \textbf{RO} \\
		\midrule
		colrow-1 & 85.88 & \textbf{82.76} & \textbf{80.76} & 69.84 & 73.44 & 74.2 \\
		\textbf{col-6} & \textbf{87.49} & 78.27 & 79.82 & \textbf{75.02} & \textbf{74.90} & \textbf{78.99} \\
		\bottomrule
	\end{tabular}
\end{table}

\begin{table*}
	\centering
	\caption{Summary of Model Design Components.}
	\vspace{-5px}
	\label{tab:model_components}
	\begin{tabular}{@{}ll@{}}
		\toprule
		\textbf{Component}                     & \textbf{Options}                                   \\ \midrule
		\textbf{Augmentation}                  & Separated Representation (Sec.~\ref{sec:backbone}), Feature Mixup (Sec.~\ref{sec:pretraining_pool}), Full Representation \\
		\textbf{Pretraining Task}              & Bag Contrastive (Sec.~\ref{sec:bag_contrastive}), Self-contrastive (Sec.~\ref{sec:pretraining_pool}), Denoising Reconstruction (Sec.~\ref{sec:pretraining_pool})  \\
		\textbf{Bag Aggregator}                & Cosine similarity (Sec.~\ref{sec:bag_contrastive}), Intersample attention mechanism (Sec.~\ref{sec:backbone}), \\
		& Cosine similarity + projection, Intersample attention mechanism + projection \\
		\textbf{Difference Contrastive Objective} & Supervised InfoNCE (Sec.~\ref{sec:diff_contrastive}), Cosine Embedding (Sec.~\ref{sec:downstream_comp}), Self-contrastive \\
		\textbf{Pseudo-pairs Generator}        & Linear Sum Assignment (Sec.~\ref{sec:diff_contrastive}), Greedy Nearest Neighbor Assignment (Sec.~\ref{sec:downstream_comp}) \\
		\textbf{Bag-level Validation Metric}   & L1, mPIoU (Sec.~\ref{sec:evaluation_methods})                                 \\ \bottomrule
	\end{tabular}
	\vspace{-5px}
\end{table*}

Our training bifurcates into pretraining and finetuning phases. The pretraining phase adopts a multi-task approach, integrating Denoising Reconstruction \cite{vincent2008extracting, somepalli2021saint}, Self-contrastive \cite{chen2020simple, bahri2021scarf, oord2018representation}, and Bag Contrastive methods. Feature Mixup \cite{somepalli2021saint, yoon2020vime, yun2019cutmix, zhang2017mixup} serves as the default pretraining augmentation for Self-contrastive. The finetuning phase's objective melds a ramping-down LLP Loss \cite{ardehaly2017co} with a ramping-up Difference Contrastive Loss. We use 50 epochs for the pretraining phase and 300 epochs for the finetuning phase, with an early stopping of 20 epochs on the validation AUC score or mPIoU score based on the real-world LLP scenario. The other optimal training hyperparameter is tuned using Optuna\footnote{\url{https://https://optuna.org/}} with 50 trials for each dataset and model. 
% The recommended configurations are presented on our GitHub repo\footnote{https://anonymous.4open.science/r/TabLLP-BDC-B7A5/}. 
We repeat 10 trials for each experiment using different train/validation/test splits. All experiments can be reproduced using a single GeForce RTX 2080 GPU.

\begin{table*}[!t]
	\caption{Coarse-grained Experimental Results.}
	\label{tab:coarse_grained_result}
	\vspace{-5px}
	\begin{tabular}{l@{\hspace{8pt}}c@{\hspace{8pt}}c@{\hspace{8pt}}c@{\hspace{8pt}}c@{\hspace{8pt}}c@{\hspace{8pt}}c@{\hspace{8pt}}|@{\hspace{8pt}}c@{\hspace{8pt}}c}
		\toprule
		& \multicolumn{6}{c}{AUC (\%)} & \multicolumn{2}{c}{Accuracy (\%)} \\
		\cmidrule(lr){2-7} \cmidrule(lr){8-9}
		& \textbf{AD} & \textbf{BA} & \textbf{CA} & \textbf{CR} & \textbf{EL} & \textbf{RO} & \textbf{Private A} & \textbf{Private B} \\
		\midrule
		Fine-grained & $87.81 \pm 0.18$ & $84.95 \pm 1.88$ & $84.85 \pm 0.93$ & $79.29 \pm 0.39$ & $77.53 \pm 1.48$ & $79.67 \pm 0.25$ & $51.39 \pm 0.16$ & $74.03 \pm 1.54$ \\
		\midrule
		DLLP & $87.22 \pm 0.65$ & $75.94 \pm 1.86$ & $76.87 \pm 1.28$ & $74.22 \pm 0.45$ & $71.37 \pm 1.58$ & $78.89 \pm 0.55$ & $46.22 \pm 0.54$ & $63.43 \pm 1.56$ \\
		LLP-GAN & $82.45 \pm 0.90$ & $70.44 \pm 0.60$ & $80.08 \pm 1.75$ & $64.44 \pm 1.78$ & $53.16 \pm 1.30$ & $70.72 \pm 0.79$ & $44.67 \pm 1.06$ & $62.76  \pm 1.48$ \\
		LLP-VAT & $83.03 \pm 1.83$ & $77.69 \pm 2.15$ & $77.25 \pm 2.32$ & $72.09 \pm 1.18$ & $72.03 \pm 2.05$ & $77.76 \pm 0.80$ & $48.49 \pm 1.15$ & $62.23 \pm 5.48$ \\
		SelfCLR-LLP & $86.14 \pm 0.77$ & $72.79 \pm 4.06$ & $79.28 \pm 0.82$ & $74.59 \pm 0.60$ & $69.59 \pm 4.00$ & $78.53 \pm 0.14$ & $48.88 \pm 0.50$ & $63.26 \pm 4.16$ \\
		\midrule
		TabLLP-DEC & $59.32 \pm 8.78$ & $55.03 \pm 3.37$ & $79.64 \pm 1.17$ & $73.79 \pm 1.18$ & $68.16 \pm 4.70$ & $62.59 \pm 1.70$ & $45.54 \pm 1.64$ & $52.21 \pm 1.96$ \\
		TabLLP-PSE & $86.15 \pm 1.01$ & $76.47 \pm 2.48$ & $75.89 \pm 1.97$ & $75.35 \pm 1.70$ & $64.93 \pm 8.55$ & $78.14 \pm 0.13$ & $43.49 \pm 2.04$ & $57.46 \pm 3.38$ \\
		TabLLP-SELF & $85.00 \pm 0.54$ & $78.22 \pm 1.14$ & $79.19 \pm 1.93$ & $75.67 \pm 0.65$ & $70.16 \pm 3.51$ & $78.86 \pm 0.31$ & $49.28 \pm 0.86$ & $56.12 \pm 1.72$ \\
		\textbf{TabLLP-BDC} & $\mathbf{87.41 \pm 0.35}$ & $\mathbf{83.81 \pm 1.67}$ & $\mathbf{80.23 \pm 1.37}$ & $\mathbf{78.49 \pm 0.25}$ & $\mathbf{73.52 \pm 0.90}$ & $\mathbf{79.34 \pm 0.32}$ & $\mathbf{50.24 \pm 1.21}$ & $\mathbf{64.99 \pm 1.29}$ \\
		\bottomrule
	\end{tabular}
	\vspace{-5px}
\end{table*}

\subsection{Baselines}
\label{sec:baselines}

Given the nascent exploration of LLP in tabular data, we also adapt several leading LLP image data baselines and introduce intuitive baselines for pure tabular scenarios. The details of selected baselines are listed below:
\begin{itemize}[leftmargin=*,nosep]
	\item {\texttt{DLLP}} \cite{ardehaly2017co}: The classical and fundamental work that proposed the LLP Loss for LLP training for image data. The design also utilizes co-trained images, text, and various label-invariant image distortions, such as rotation and flipping. We only keep the LLP Loss because our task scenario does not satisfy multimodal data and label-invariant augmentation. We replace the Xception (for image input) and MLP (for text input) backbones with a single SOTA supervised deep tabular learning model \cite{somepalli2021saint} we adopted in the study for a fair comparison. DLLP stands as a primary comparison model in our ablation study, highlighting the enhanced performance offered by the Difference Contrastive objective.
	\item {\texttt{LLP-GAN}} \cite{liu2022llp}: A classical work that leverages the power of  SGAN \cite{odena2016semi} to LLP. The architecture consists of a generator and a discriminator (K+1-way classifier). It combines the traditional GAN loss with LLP Loss \cite{ardehaly2017co} to encourage the predicted label proportions to match the true label proportions. The work is proposed for image data. To tackle the challenge of generating categorical features and realistic tabular samples, we adopt the design of cWGAN \cite{engelmann2021conditional} as the backbone for the generator and discriminator while the training objective and procedure remain consistent with LLP-GAN.
	\item {\texttt{LLP-VAT}} \cite{tsai2020learning}: A notable work that applies consistency regularization upon the LLP Loss \cite{ardehaly2017co} with the help of Virtual Adversarial Training \cite{miyato2018virtual}. The model will learn to be robust to adversarial perturbations, which can improve its generalization ability. The work is proposed for image data but can be directly transferred to tabular data. The backbone encoder is upgraded to the same model \cite{somepalli2021saint} we used in our model architecture.
	\item {\texttt{SelfCLR-LLP}} \cite{nandy2022domain}: To the best of our knowledge, it is the only work that tackles the LLP problem in tabular data. The model combines the Self-contrastive loss with the LLP Loss \cite{ardehaly2017co} to promote controlled diversity in the representations. The original backbone was AutoInt \cite{song2019autoint}, and we replaced it with a more recent model \cite{somepalli2021saint}, as in our general model setting, for fairness. SelfCLR-LLP stands as a primary comparison model in our ablation study, highlighting the enhanced performance offered by the Difference Contrastive objective.
	\item {\texttt{TabLLP-DEC}}: DEC \cite{xie2016unsupervised} is a typical deep clustering model in an unsupervised data setting. The model continuously updates the embeddings of the centroids during the training process by fitting the student's t-distribution. An intuitive modification to transfer it to the LLP setting as a baseline is to add the LLP Loss \cite{ardehaly2017co} to its clustering loss. The encoder is the same as the other models \cite{somepalli2021saint} in the experiment.
	\item {\texttt{TabLLP-PSE}}: Pseudo-labeling is another typical approach \cite{liu2021two, caron2018deep} to further finetune bag-level training with instance-level pseudo-labels. We implement constrained KMeans to generate pseudo-labels that exactly match the bag label proportion and perform supervised learning with respect to pseudo-labels in the second stage after initial training with the LLP Loss \cite{ardehaly2017co}. The encoder is the same as the other models \cite{somepalli2021saint} in the experiment.
	\item {\texttt{TabLLP-SELF}}: This variant draws inspiration from SelfCLR-LLP \cite{nandy2022domain}. However, it diverges by employing traditional Self-contrastive learning \cite{somepalli2021saint, bahri2021scarf} with feature mixup augmentation \cite{somepalli2021saint, yoon2020vime, yun2019cutmix, zhang2017mixup} as its auxiliary loss. The encoder configuration aligns with that of other baselines \cite{somepalli2021saint}. TabLLP-Self stands as a primary comparison model in our ablation study, highlighting the enhanced performance offered by the Difference Contrastive objective.
\end{itemize}

\begin{table}[!t]
	\caption{Relationship between mPIoU and AUC Metrics.}
	\label{tab:coarse_grained_saint}
	\vspace{-5px}
	\begin{tabular}{llccccc}
		\toprule
		& & \multicolumn{5}{c}{Dataset} \\
		\cmidrule(lr){3-7}
		\textbf{Model} & \textbf{Metric} & \textbf{AD} & \textbf{BA} & \textbf{CA} & \textbf{CR} & \textbf{EL} \\
		\midrule
		Supervised & mPIoU & 87.76 & \textbf{90.23} & \textbf{97.90} & 87.05 & \textbf{95.14} \\
		Supervised & AUC & \textbf{91.37} & \textbf{94.01} & \textbf{95.87} & \textbf{82.36} & \textbf{90.73} \\
		\midrule
		DLLP & mPIoU & 94.55 & 78.59 & 96.01 & \textbf{93.17} & 91.96 \\
		DLLP & AUC & 87.22 & 75.94 & 76.87 & 74.22 & 71.37 \\
		\midrule
		TabLLP-BDC & mPIoU & \textbf{94.96} & 78.78 & 95.72 & 93.09 & 94.92 \\
		TabLLP-BDC & AUC & 87.41 & 83.81 & 80.23 & 78.49 & 73.52 \\
		\bottomrule
	\end{tabular}
	\vspace{-10px}
\end{table}

\section{Experimental Results}
\label{sec:coarse_grained}
We present the remaining experimental results as follow:
\begin{itemize}[leftmargin=*,nosep]
	\item The hidden instance-level test performance of the coarse-grained validation scenario is reported in Tab.~\ref{tab:coarse_grained_result}. Our TabLLP-BDC achieved SOTA performance in the practical scenario devoid of fine-grained validation.
	\item Relationship between mPIoU and AUC metrics can be observed in Tab.~\ref{tab:coarse_grained_saint}. The imperfect correlation between the best instance-level predictions and label proportions validates our hypothesis that solely relying on bag-level supervision is insufficient for accurate instance-level prediction in tabular data.
	% \item In order to compare the performance with SelfCLR-LLP \cite{nandy2022domain} in more detail, we also conducted comparative experiments on its item recommendation dataset Movielens-1M \cite{harper2015movielens}. As shown in Tab.~\ref{tab:movie}, our TabLLP-BDC exhibits superiority over other LLP baselines, including SelfCLR-LLP.
\end{itemize}

% \begin{table}[H]
% 	\caption{Fine-grained Experimental Results on Movielens-1M.}
% 	\label{tab:movie}
% 	\vspace{-5px}
% 	\centering
% 	\begin{tabular}{l|c}
% 		\hline
% 		\textbf{Model} & \textbf{AUC} \\
% 		\hline
% 		Supervised & $86.02 \pm 0.1$ \\
% 		\hline
% 		DLLP & $73.75 \pm 0.81$ \\
% 		LLP-GAN & $74.31 \pm 1.52$ \\
% 		LLP-VAT & $71.12 \pm 1.20$ \\
% 		SelfCLR-LLP & $74.36 \pm 1.31$ \\
% 		TabLLP-DEC & $55.34 \pm 0.50$ \\
% 		TabLLP-PSE & $74.18 \pm 0.36$ \\
% 		TabLLP-SELF & $74.10 \pm 0.79$ \\
% 		\textbf{TabLLP-BDC} & $\mathbf{75.06 \pm 0.58}$ \\
% 		\hline
% 	\end{tabular}
% 	\vspace{-5px}
% \end{table}

\section{Ablation Study}
\label{sec:ablation}
We present a comprehensive ablation study to assess the individual and combined effectiveness of the components, as summarized in Tab.~\ref{tab:model_components}, within our TabLLP-BDC model, including the studies of pretraining tasks, augmentation technique, bag aggregator, downstream contrastive objectives, pseudo-pairs generator, and mPIoU evaluation metric. For consistency, all experiments are conducted with a fixed bag size of 256 across five medium-sized public datasets, each with accessible fine-grained validation. Hyperparameters are meticulously tuned for each study.

\subsection{Ablation Study of Downstream Contrastive Objectives}
\label{sec:downstream_obj}

\begin{table}[!t]
	\caption{Ablation Study of Downstream Contrastive Objectives.}
	\label{tab:downstream}
	\vspace{-5px}
	\begin{tabular}{llccccc}
		\toprule
		& & \multicolumn{5}{c}{AUC (\%)} \\
		\cmidrule(lr){3-7}
		\textbf{Pretrain} & \textbf{Model} & \textbf{AD} & \textbf{BA} & \textbf{CA} & \textbf{CR} & \textbf{EL} \\
		\midrule
		No & DLLP & \textbf{87.64} & 77.18 & 78.75 & 75.07 & 74.33 \\
		No & SelfCLR & 87.44 & 79.73 & 81.34 & 75.43 & 74.44 \\
		No & SELF & 86.66 & 79.38 & \textbf{82.55} & 77.04 & 74.87 \\
		No & BDC & 87.03 & \textbf{81.77} & 80.64 & \textbf{79.10} & \textbf{75.43} \\
		\midrule
		Optimal & DLLP & 87.76 & 84.36 & 84.40 & 75.60 & 75.64 \\
		Optimal & SelfCLR & 87.67 & 83.50 & 84.78 & 76.13 & 75.40 \\
		Optimal & SELF & 86.99 & 83.16 & 84.19 & 77.69 & 75.44 \\
		Optimal & BDC & \textbf{87.81} & \textbf{84.95} & \textbf{84.85} & \textbf{79.29} & \textbf{77.53} \\
		\bottomrule
	\end{tabular}
	\vspace{-5px}
\end{table}

In this section, we evaluate the performance of different finetuning objectives, ensuring the pretraining task remains constant. Specifically, our focus is on the following contrastive-based finetuning objectives: LLP loss \cite{ardehaly2017co}, the loss proposed in SelfCLR-LLP \cite{nandy2022domain}, LLP loss + Self-contrastive \cite{chen2020simple} (similar to SelfCLR-LLP \cite{nandy2022domain} but using Feature Mixup augmentation \cite{zhang2017mixup, yun2019cutmix, yoon2020vime, somepalli2021saint}), and LLP loss + Difference Contrastive. When evaluating each finetuning objective, we consider two distinct pretraining scenarios: using the best pretraining tasks combination identified for each finetuning objective from our previous experiments in Tab.~\ref{tab:pretraining}, and a scenario without any pretraining. As shown in Tab.~\ref{tab:downstream}, Our Difference Contrastive model shows superior performance compared to all the other contrastive-based objectives in most scenarios and datasets. To highlight, when combined with the best pretraining task for each objective, the best performance is always achieved by our Difference Contrastive objectives.

\subsection{Ablation Study of Downstream Components}
\label{sec:downstream_comp}

\begin{table}[!t]
	\caption{Ablation Study of Pseudo-pairs Generator.}
	\label{tab:generator}
	\vspace{-5px}
	\begin{tabular}{lccccc}
		\toprule
		& \multicolumn{5}{c}{AUC (\%)} \\
		\cmidrule(lr){2-6}
		\textbf{Assignment} & \textbf{AD} & \textbf{BA} & \textbf{CA} & \textbf{CR} & \textbf{EL} \\
		\midrule
		Greedy & 87.57 & 84.46 & \textbf{85.31} & \textbf{79.66} & 76.01 \\
		\textbf{Linear} & \textbf{87.81} & \textbf{84.95} & 84.85 & 79.29 & \textbf{77.53} \\
		\bottomrule
	\end{tabular}
	\vspace{-5px}
\end{table}

\begin{table}[!t]
	\caption{Ablation Study of Difference Contrastive Loss.}
	\label{tab:loss}
	\vspace{-5px}
	\begin{tabular}{lccccc}
		\toprule
		& \multicolumn{5}{c}{AUC (\%)} \\
		\cmidrule(lr){2-6}
		\textbf{Loss Function} & \textbf{AD} & \textbf{BA} & \textbf{CA} & \textbf{CR} & \textbf{EL} \\
		\midrule
		CosineEmbedding & \textbf{87.81} & 83.40 & \textbf{84.85} & 75.84 & 75.41 \\
		\textbf{SupInfoNCE} & 87.33 & \textbf{84.95} & 84.43 & \textbf{79.29} & \textbf{77.53} \\
		\bottomrule
	\end{tabular}
	\vspace{-10px}
\end{table}

To further refine the generation of pseudo-positive pairs, we introduce two methodologies: Linear Sum Assignment \cite{crouse2016implementing} and Greedy Nearest Neighbor Assignment. Additionally, we explore two learning objectives: the Supervised InfoNCE-based Contrastive loss and the Cosine Embedding loss\footnote{\url{https://pytorch.org/docs/stable/generated/torch.nn.CosineEmbeddingLoss.html}}. The Greedy Nearest Neighbor Assignment will iterate through the similarity matrix to greedily extract the highest similarity 1-to-1 pairs to generate pseudo-positive pairs. The Cosine Embedding loss will treat the remaining pairs as pseudo-negative pairs and encourage the two sample representations to be similar or dissimilar based on both positive and negative pairs. Tab.~\ref{tab:generator} and Tab.~\ref{tab:loss} show that the Linear Sum Assignment and the Supervised InfoNCE-based Contrastive loss have empirically better performance. Please note that while the performance gap between the two generators is small, the Linear Sum Assignment has a significant efficiency advantage over the Greedy Nearest Neighbor Assignment in the implementation, which is crucial for operations that run every mini-batch. Therefore, we recommend using the Linear Sum Assignment for a practical balance between effectiveness and efficiency.

\subsection{Ablation Study of Pretraining Tasks}
\label{sec:pretraining_task}
In this section, we evaluate the performance of various pretraining tasks, ensuring the finetuning objective remains constant throughout. Following the standard pretraining methodology outlined in \cite{somepalli2021saint} and the Sec.~\ref{sec:pretraining_pool}, we consistently employ the denoising reconstruction task. Our primary focus is on the contrastive task, where we juxtapose the baseline Self-contrastive \cite{somepalli2021saint, bahri2021scarf} approach with our proposed Bag Contrastive method. The results from this comparison in Tab.~\ref{tab:pretraining} shed light on the impact of different pretraining contrastive tasks under three distinct finetuning objectives: the LLP loss \cite{ardehaly2017co}, the loss proposed in SelfCLR-LLP \cite{nandy2022domain}, LLP loss + Self-contrastive \cite{chen2020simple} (similar to SelfCLR-LLP \cite{nandy2022domain} but using Feature Mixup augmentation \cite{zhang2017mixup, yun2019cutmix, yoon2020vime, somepalli2021saint}), and LLP loss + Difference Contrastive. Notably:
\begin{itemize}[leftmargin=*,nosep]
	\item The inclusion of a pretraining phase consistently enhances performance, emphasizing its critical role in label proportion learning from tabular data.
	\item Our Bag Contrastive pretraining emerges as a pivotal component, effectively bridging the gap between instance-level pretraining and bag-level finetuning.
	\item The introduction of the Difference Contrastive sometimes diminishes the distinct advantages of our Bag Contrastive, suggesting potential overlaps or redundancies between the two.
\end{itemize}
This study underscores a key insight: our Bag Contrastive and Difference Contrastive, which emphasize bag-level pretraining and instance-level supervision respectively, are designed to complement and enhance the instance-level pretraining offered by the Self-contrastive method \cite{somepalli2021saint, bahri2021scarf} and the bag-level supervision provided by the LLP loss \cite{ardehaly2017co} respectively. Their major role is to bolster the consistency between learning and prediction. However, since they both narrow the gap of LLP difficulty toward each other, they do not seem to significantly amplify each other's benefits together.

\subsection{Ablation Study of Pretraining Components}
\label{sec:pretraining_comp}

\begin{table}[!t]
	\caption{Ablation Study of Augmentation.}
	\label{tab:augmentation}
	\vspace{-5px}
	\begin{tabular}{llccccc}
		\toprule
		& & \multicolumn{5}{c}{AUC (\%)} \\
		\cmidrule(lr){3-7}
		\textbf{Model} & \textbf{Augment.} & \textbf{AD} & \textbf{BA} & \textbf{CA} & \textbf{CR} & \textbf{EL} \\
		\midrule
		DLLP & FM & 87.46 & 81.43 & \textbf{83.53} & 75.52 & \textbf{75.31} \\
		DLLP & SR & \textbf{87.55} & \textbf{84.36} & 82.30 & \textbf{75.58} & 75.24 \\
		\midrule
		BDC & FM & \textbf{87.79} & 82.18 & \textbf{84.85} & \textbf{79.29} & \textbf{76.24} \\
		BDC & SR & \textbf{87.79} & \textbf{84.95} & 81.85 & 79.18 & \textbf{76.24} \\
		\bottomrule
	\end{tabular}
	\vspace{-10px}
\end{table}

Furthermore, our study delves into two essential components of the pretraining task: data augmentation and bag aggregation.

Given the inherent challenges of generating label-invariant data augmentation for tabular data, we introduce an augmentation-free approach that uses bag comparison instead. However, by leveraging the capabilities of the CLS token in the Tabular Transformer \cite{gorishniy2021revisiting, somepalli2021saint}, we also propose to adopt representation separation to extract two label-invariant views of samples directly. This approach mirrors the idea of subsetting raw features in SubTab \cite{ucar2021subtab} but is executed in the representation space using the CLS token and basic representation space. Our experiments in Tab.~\ref{tab:augmentation}, comparing this separated representation (SR) technique with the classical feature mixup (FM) method \cite{zhang2017mixup, yun2019cutmix, yoon2020vime, somepalli2021saint}, reveal varied performance across datasets. However, our separated representation offers a slight efficiency advantage as it avoids direct sample manipulation.

In our exploration of aggregation techniques, we compare the performance of a similarity-driven weighted sum aggregator (WS) with an intersample attention-based aggregator (IA). The former calculates attention scores based on the cosine similarity between samples from the same bag, contributing to the overall bag representation. In contrast, the latter, inspired by the design of multi-head row attention in SAINT \cite{somepalli2021saint} and Sec.~\ref{sec:backbone}, aggregates intersample representations into a singular bag representation vector using mean reduction. Our experiments shown in Tab.~\ref{tab:aggregation} indicate that the similarity-driven weighted sum aggregator, when used without projection, generally delivers superior performance.

\begin{table}[!t]
	\caption{Ablation Study of Bag Aggregation Techniques.}
	\label{tab:aggregation}
	\vspace{-5px}
	\begin{tabular}{llccccc}
		\toprule
		& & \multicolumn{5}{c}{AUC (\%)} \\
		\cmidrule(lr){3-7}
		\textbf{Model} & \textbf{Augment.} & \textbf{AD} & \textbf{BA} & \textbf{CA} & \textbf{CR} & \textbf{EL} \\
		\midrule
		BDC & WS & 87.74 & \textbf{84.91} & 83.55 & \textbf{79.16} & \textbf{77.53} \\
		BDC & IA & \textbf{87.81} & 84.42 & 83.75 & 79.00 & 77.22 \\
		BDC & WS-Proj & 87.64 & 84.11 & \textbf{83.94} & 78.72 & 76.37 \\
		BDC & IA-Proj & 87.65 & 83.85 & 80.93 & 79.13 & 76.45 \\
		\bottomrule
	\end{tabular}
	\vspace{-5px}
\end{table}

\subsection{Evaluation of Bag-Level Metrics in Early Stopped LLP Models}

\begin{table}[!t]
	\caption{Ablation Study of Bag-Level Metrics in Early Stopped LLP Models.}
	\label{tab:metrics}
	\vspace{-5px}
	\begin{tabular}{llccccc}
		\toprule
		& & \multicolumn{5}{c}{AUC (\%)} \\
		\cmidrule(lr){3-7}
		\textbf{Model} & \textbf{Metrics} & \textbf{AD} & \textbf{BA} & \textbf{CA} & \textbf{CR} & \textbf{EL} \\
		\midrule
		van. DLLP & L1 & \textbf{87.22} & \textbf{75.94} & 74.42 & 73.50 & 69.68 \\
		van. DLLP & \textbf{mPIoU} & 86.62 & 74.20 & \textbf{76.87} & \textbf{74.22} & \textbf{71.37} \\
		\midrule
		DLLP & L1 & \textbf{87.32} & 77.84 & 78.45 & \textbf{74.69} & 70.00 \\
		DLLP & \textbf{mPIoU} & 87.30 & \textbf{80.74} & \textbf{78.85} & 74.03 & \textbf{72.29} \\
		\midrule
		SELF & L1 & 84.38 & \textbf{80.01} & \textbf{80.72} & 75.57 & 72.95 \\
		SELF & \textbf{mPIoU} & \textbf{85.13} & 79.96 & 80.46 & \textbf{76.43} & \textbf{73.29} \\
		\midrule
		BDC & L1 & 86.80 & 83.14 & 78.64 & 78.25 & 70.35 \\
		BDC & \textbf{mPIoU} & \textbf{87.41} & \textbf{83.81} & \textbf{80.23} & \textbf{78.49} & \textbf{73.52} \\
		\bottomrule
	\end{tabular}
	\vspace{-10px}
\end{table}

\label{sec:bag_metrics}
We investigate the performance of LLP models that are early stopped using bag-level evaluation metrics. Specifically, we employ the L1 metric \cite{tsai2020learning} and our mPIoU metric (Sec.~\ref{sec:evaluation_methods}). The context for this examination parallels the scenarios outlined in the Downstream Contrastive Objectives ablation study: using the best pretraining tasks combination except for the vanilla DLLP. Tab.~\ref{tab:metrics} reveals that our mPIoU metric offers more counterpoint advantage over the L1. However, it is important to note that despite numerous experiments, some degree of fluctuation persists in the outcomes. Since this particular aspect is not the central focus of our study, we have incorporated both metrics in our prior Performance Evaluation of Coarse-grained Validation, selecting the highest results for presentation in Tab.~\ref{tab:coarse_grained_result}. For future research, we advocate the consideration of both these metrics to ensure a comprehensive evaluation.

\end{document}